\documentclass[pdflatex,sn-nature]{sn-jnl}

\usepackage[english]{babel}
\usepackage{overpic}
\usepackage{graphicx}
\usepackage{amsmath,amssymb}
\usepackage{booktabs}
\usepackage{tabularx}
\usepackage{array}
\usepackage{ragged2e}
\usepackage{pifont}
\usepackage{float}
\usepackage{tcolorbox}
\usepackage[inline]{enumitem}
\usepackage{overpic}
\usepackage{tikz}
\usepackage{xcolor}
\usepackage{rotating}      
\hypersetup{colorlinks=true, linkcolor=black, citecolor=black, urlcolor=black}

\usetikzlibrary{
  shadows,
  shapes.misc,
  shapes.geometric,
  arrows.meta,
  positioning,
  decorations.pathmorphing,
  calc,
  fit
}

\newcolumntype{Y}[1]
  {>{\hsize=#1\hsize\RaggedRight\arraybackslash}X}
\newcolumntype{J}[1]
  {>{\hsize=#1\hsize\justifying\arraybackslash}X}

\newcommand{\xmark}{\ding{55}}

\definecolor{revthreecolor}{rgb}{0.00,0.20,0.75}
\newcommand{\revthree}[1]{\textcolor{black}{#1}}

\begin{document}

\title{Safe Learning for Contact-Rich Robot Tasks: A Survey from Classical
Learning-Based Methods to Safe Foundation Models}

\author*[1,2,3]{\fnm{Heng} \sur{Zhang}}\email{heng.zhang@iit.it}
\author[4]{\fnm{Rui} \sur{Dai}}
\author[1]{\fnm{Gokhan} \sur{Solak}}
\author[3]{\fnm{Pokuang} \sur{Zhou}}
\author[4]{\fnm{Nikos} \sur{Tsagarakis}}
\author[3,5,6]{\fnm{Yu} \sur{She}}
\author[1]{\fnm{Arash} \sur{Ajoudani}}

\affil[1]{\orgdiv{Human-Robot Interfaces and Interaction Lab},
  \orgname{Istituto Italiano di Tecnologia}, \city{Genova}, \country{Italy}}
\affil[2]{\orgdiv{Ph.D. program of national interest in Robotics and
  Intelligent Machines (DRIM)}, \orgname{Universit\`a di Genova},
  \city{Genoa}, \country{Italy}}
\affil[3]{\orgdiv{Edwardson School of Industrial Engineering},
  \orgname{Purdue University}, \city{West Lafayette}, \state{IN},
  \country{USA}}
\affil[4]{\orgdiv{Humanoids and Human Centered Mechatronics Lab}, \orgname{Istituto Italiano di Tecnologia},
  \city{Genova}, \country{Italy}}
\affil[5]{\orgdiv{School of Mechanical Engineering},
  \orgname{Purdue University}, \city{West Lafayette}, \state{IN},
  \country{USA}}
\affil[6]{\orgdiv{Elmore Family School of Electrical and Computer
  Engineering}, \orgname{Purdue University}, \city{West Lafayette},
  \state{IN}, \country{USA}}

\abstract{Contact-rich tasks remain difficult for robotic systems due to
uncertainty, complex dynamics, and the risk of damage during interaction. As
learning-based methods increasingly enable robots to acquire and generalize
advanced manipulation skills, ensuring safety during both learning and
deployment has become essential. We present, to our knowledge, the first
scoping survey that places safety at the center of learning for contact-rich
robotic tasks, spanning classical provable safety, safe reinforcement
learning, and emerging robotic foundation models. We analyze how safety
principles are formulated, enforced, and evaluated across established learning
approaches and foundation-model-enabled methods. Existing work is organized
into two domains: safe exploration, which reduces unsafe actions during
learning, and safe execution, which ensures robustness and constraint
satisfaction during deployment. We further examine how safe learning
principles extend to robotic foundation models, particularly vision-language
models (VLMs) and vision-language-action models (VLAs). These models introduce
opportunities such as language-based constraint specification and multimodal
safety grounding, while also posing new risks and evaluation challenges.
Finally, we outline current limitations and highlight promising directions
toward reliable, safety-aligned robotic systems capable of operating in
complex, contact-rich environments.}


\maketitle


\begin{quote}
    \textit{``Contact is the heart of robotic manipulation. To understand manipulation, you must understand contact."}\\
    \href{https://mtmason.com/the-heart-of-robotic-manipulation/}{The Heart of Robotic Manipulation}\\
    \hfill \textbf{------ Matthew T. Mason}
\end{quote}
\section{Introduction} \label{sec:intro}
Robots are increasingly expected to operate in unstructured and dynamic environments, where interaction with the physical world is inevitable~\citep{billard2019trends, ajoudani2018progress}. Among such interactions, contact-rich tasks such as assembly, insertion, cutting, or tool use are particularly challenging due to their complex, discontinuous dynamics and the need for precise force and motion control. Traditional model-based control methods often struggle in these scenarios due to inaccurate or hard-to-obtain models, while learning-based methods promise flexibility and adaptability by leveraging data-driven representations~\citep{elguea2023review}.

However, applying learning-based methods to contact-rich tasks raises critical safety concerns. Exploration during training may lead to unsafe actions that damage the robot, the environment, or nearby humans. Similarly, the execution of learned policies in deployment must guarantee constraint satisfaction and robustness, especially under distributional shifts. Therefore, safety must be addressed both during exploration and during execution to ensure the reliable operation of autonomous systems in the real world.

This survey focuses on safe learning-based approaches that address the dual challenges of safe exploration and safe execution in robot contact-rich tasks. We categorize and analyze a wide range of methods, including constrained and risk-aware reinforcement learning, uncertainty-aware control, model-based safety filters, and hybrid approaches that combine learning with formal guarantees. These methods aim to strike a balance between learning efficiency, task performance, and safety assurance.

We begin by outlining the unique challenges posed by contact-rich tasks and the requirements for safety in this context. We then present a taxonomy of existing methods based on the learning paradigm and safety mechanism employed. Finally, we identify open problems and future research directions, including benchmark design, generalization, and integration of safety across system levels.

This survey aims to provide a structured overview of the current landscape, foster a deeper understanding of safe learning in contact-rich robotics, and inspire new research toward safe and capable autonomous systems. In particular, it emphasizes the emerging role of vision-language models (VLMs) and vision-language-action models (VLAs) as robotic foundation models that unify perception, language, vision, and control, but {safety} in contact-rich tasks is still an open challenge~\citep{330, zhang2026compliantvla}. By framing safety challenges and design principles in the context of these models, the survey seeks to provide timely guidance for the development of VLM/VLA-based systems that are not only powerful and general but also verifiably safe in real-world deployment.

A number of recent surveys have explored topics at the intersection of safety, learning, and contact-rich robotic manipulation. However, none comprehensively address the unique challenges and opportunities posed by safe learning in contact-rich tasks. Our survey aims to bridge this gap by providing a focused, methodical overview of the state-of-the-art, specifically within this critical niche.

Several surveys have investigated safe learning and reinforcement learning (RL) more generally. {An early and influential reference is the comprehensive safe-RL survey of García and Fernández~\citep{garcia2015comprehensive}, whose taxonomy remains a common organizing lens for subsequent work, separating approaches that modify the optimization criterion from those that modify the exploration process.} Works such as~\citep{brunke2022safe, zhao2023state, gu2024review} provide comprehensive overviews of safety mechanisms in learning-based control and safe RL, including constraints, shielding, and exploration strategies. More specialized perspectives are offered in surveys on state-wise safety~\citep{zhao2023state}, control barrier functions in RL~\citep{guerrier2024learning} and shielded RL~\citep{odriozola2023shielded}. While these surveys address safety in learning, they often neglect the complexities introduced by physical interaction and contact dynamics.

From the perspective of robotic manipulation and contact, surveys such as~\citep{suomalainen2022survey, kroemer2021review} and ~\citep{xu2019compare} analyze manipulation strategies and control frameworks, including both model-based and model-free methods. Others review variable impedance control~\citep{abu2020variable} and compliant mechanisms~\citep{samadikhoshkho2025review}, which are crucial for adapting to contact uncertainties but do not directly integrate learning-based safety frameworks.

The application of RL in robotic assembly has also been reviewed in recent literature~\citep{stan2020reinforcement, elguea2023review, das2025towards} with emphasis on task success and sample efficiency. However, these surveys primarily focus on task performance and learning architectures, offering limited discussion on safety guarantees during exploration or deployment. 
A recent survey paper \citep{Tsuji2025asu} reviews the imitation learning methods for contact-rich robotic tasks. 
Differently, we center our review specifically on online safe learning methods, without focusing on imitation learning.

More recent efforts begin to address physical risk and safety in the era of data-driven and foundation model-based robotics~\citep{kojima2025comprehensive, lamba2025alignment, xie2025towards, wabersich2023data},  highlighting the growing relevance of scalable safety mechanisms. However, they often center on high-level policy alignment or general-purpose systems, lacking specificity in contact-rich manipulation contexts.

This survey also concentrates on the key safety gaps and data requirements underlying robot foundation models~\cite{330, 33}. Advanced agentic embodied AI sheds light on a generalist robotic system, while the safety issue remains due to a lack of physical interaction data~\cite{salimpour2025towards}. 
Safety for VLA robotic generalists lags because high-fidelity, failure and near‑miss contact data are scarce, and foundational semantic models still mis‑ground force, region, and tolerance constraints in physically complex interaction regimes \citep{cruz2024learning}.  
Hybrid plan–parameterize–filter stacks that couple semantic/VLM layers with compliant action abstractions and certified online shields improve robustness under distribution shift but remain challenged by contact mode transitions and geometry/material variation \citep{33, 330, 10705419, das2025towards}.
Advancing VLA safety thus hinges on curated multimodal contact corpora, standardized force/violation metrics, and tighter analytical–data-driven integration for adaptive constraint grounding and reactive invariance maintenance \citep{salimpour2025towards}.

{Most closely related to the present work is the concurrent review of~\citep{parnada2025towards}, which surveys cost-effective and safe contact-rich robotic manipulation with reinforcement learning. Our scope is broader and complementary: rather than restricting to RL, we (i) place safety across both exploration and execution at the center specifically for contact-rich interaction; (ii) bridge classical provable-safety methods (Lyapunov, control barrier functions, contraction metrics, and reachability) with learning-based and robotic foundation-model (VLM/VLA) approaches under a single set of safety axes; and (iii) contribute a multi-perspective categorization spanning sensing/policy modality, safety enforcement space, safety abstraction level, a data pyramid, and a three-level evaluation scheme. Accordingly, rather than claiming priority in absolute terms, we position this survey as the first to jointly span these classical and foundation-model perspectives for safe contact-rich learning.}

Lastly, surveys on human-like manipulation~\citep{huang2025human}, force/torque control~\citep{sy2025review},  and safety in robotic manufacturing~\citep{lakshminarayanan2024robots} provide valuable domain-specific insights but remain largely disconnected from the safe learning perspective, although safety standards are discussed in~\citep{lakshminarayanan2024robots}.

\begin{figure}
    \centering
    \begin{overpic}[width=0.90\linewidth]{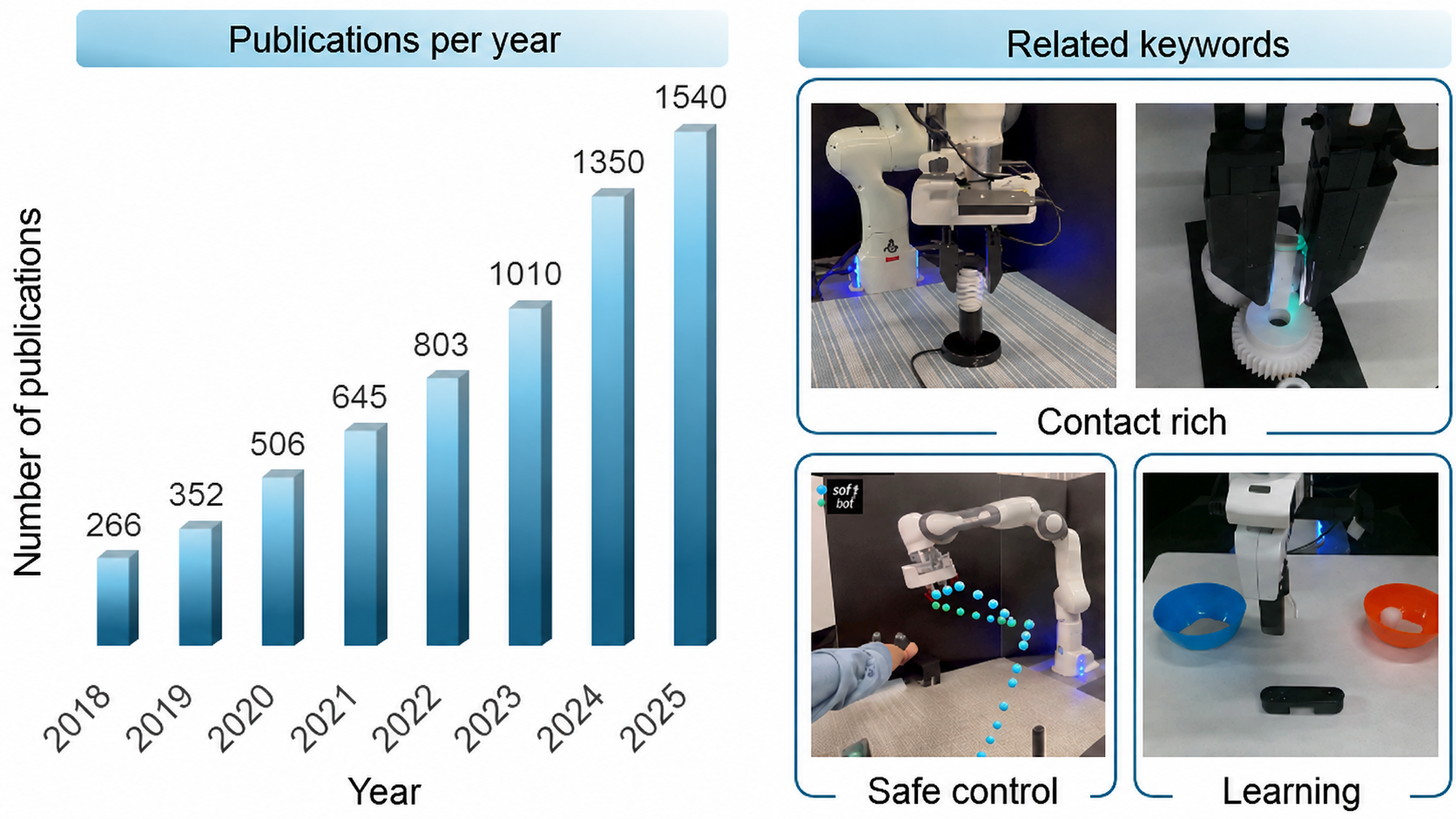}
    \put(46,4){\footnotesize\rotatebox{52}{(partial)}}
    \end{overpic}
    \caption{{Trend of research publications that include keywords related to both safety, contact-rich robotics and learning. Representative example works are shown: contact-rich tasks, safe control, and learning-based methods \cite{11023035, luu2025manifeel}.}
    \revthree{The count for 2025 is incomplete: it contains only the publications indexed up to our literature-search cut-off (August 2025), so the 2025 bar is not directly comparable with the complete years. Year-over-year comparisons should therefore be based on the complete years 2018--2024, and the trend should be read from the yearly counts rather than from a single overall growth percentage.}
    }
    \label{fig:trend} 
\end{figure}

In contrast to the aforementioned surveys, our work {focuses specifically on} the intersection of safe learning and contact-rich robotics, where safety must be ensured not only in policy learning but also in dynamic interaction with the environment. We systematically review methods that account for safety during physical contact, discuss open challenges in real-world applications, and outline promising directions for future research. We conducted a detailed analysis of the research intensity in this field over recent years, covering the period from 2018 through 2025. {Because our literature search has a fixed cut-off date, the 2025 count is necessarily partial; we therefore mark 2025 as incomplete and base any year-over-year comparison on the complete years (2018--2024) rather than on the partial 2025 data.} Figure~\ref{fig:trend} illustrates the trend of research interest in this area, along with the examples of robotic experiments associated with the keywords “safety,” “contact-rich robotics,” and “learning.”

Therefore, it is necessary to provide a comprehensive and timely survey of recent advancements in safe learning for robotic contact-rich tasks. 
This survey focuses on learning-based approaches that directly integrate safety mechanisms into the learning process, such as safe reinforcement learning and model-based safety strategies. 
{To be precise, we exclude \emph{pure} imitation-learning methods without an online mechanism for actively managing safety during exploration or execution. We do \emph{not}, however, exclude every method that uses demonstration data: approaches that are pre-trained or bootstrapped from demonstrations, including VLA policies, fall within our scope when they additionally incorporate an online-learning component and/or an explicit safety mechanism (constraint satisfaction, shielding, compliant enforcement, or risk-aware objectives). 
It is this criterion, rather than the learning-paradigm label, that determines inclusion.} 
{Our scope of contact-rich tasks is defined explicitly in Section of Robotic Contact-Rich Tasks}

In light of the aforementioned challenges and recent advancements, we systematically review the landscape of safe learning methods for robotic contact-rich tasks, which pose unique challenges due to the dynamic, force-intensive, and uncertain nature of physical interactions. Our contributions are threefold:
\begin{enumerate}
    \item  We introduce a structured safety-centric taxonomy that categorizes safe learning approaches by key dimensions, including learning phase (exploration vs. execution), level of safety integration (planning, control, or end-to-end), and modalities used (force/torque, vision, etc.). This survey provides a comprehensive lens for analyzing existing methods and identifying safety design trade-offs.
    \item  Beyond general safe learning, we focus on its application to contact-rich robotic tasks such as insertion, polishing, and assembly. We detail how safety constraints are embedded in these tasks, and map the methods used to specific operational challenges (e.g., compliance, {contact-invariant} tasks, collision avoidance, and force control).
    \item  We synthesize open research questions and outline critical challenges such as sim-to-real transfer under safety constraints, the scarcity of standardized benchmarks, and the need for provably safe generalization. We also discuss underexplored directions, including hybrid control-learning frameworks and human-in-the-loop safety mechanisms. Most importantly, we highlight the challenges and future opportunities in integrating safe contact-rich learning with large robotic foundation models, particularly VLM and VLA.
\end{enumerate}


\section{Results}
This section presents the main outcomes of our scoping review through a structured synthesis of the existing literature on safe learning for contact-rich robotic tasks. The findings of our review are organized to highlight the current state of the field, including the principal technical challenges, methodological trends, representative approaches, and emerging research directions. By systematically analyzing and categorizing the reviewed studies, we identify common design principles, unresolved limitations, and opportunities for future development, providing a comprehensive overview of the evolving landscape of safe learning for contact-rich manipulation.

\subsection{Categorization of Methods from Different Perspectives} \label{sec:methods}
This section organizes existing safe learning approaches for contact-rich robot tasks along several complementary dimensions to clarify their underlying assumptions and design trade-offs. By examining methods through perspectives such as task characteristics, sensing and policy modalities, data acquisition strategies, simulation environments and benchmarks, safety evaluation metrics, safety abstraction levels, and safety enforcement spaces, the survey highlights common patterns and gaps that are not apparent from any single taxonomy. This multi-view categorization also helps connect classical control-theoretic techniques with modern learning-based and foundation-model-based approaches, providing a unified lens for analyzing current methods and guiding future developments.

\begin{figure}[t]
    \centering
    \begin{tikzpicture}[
      font=\footnotesize, every node/.style={align=center},
      s/.style={draw=gray, line width=0.7pt},
      f/.style={-{Latex}, draw=black, line width=0.4pt},
      lbl/.style={text=black, font=\footnotesize}
    ]
    \begin{scope}[shift={(0,0)}]
      \draw[s, fill=teal!8] (0,0) rectangle (3.4,0.35);
      \draw[s, fill=magenta!18] (1.3,0.35) rectangle (2.1,1.05);
      \draw[f] (1.7,1.55) -- (1.7,1.08) node[midway,right,lbl]{$F_n$};
      \draw[f] (2.25,0.7) -- (3.0,0.7); \draw[f] (1.15,0.7) -- (0.4,0.7);
      \node[lbl] at (1.7,-0.45) {Surface grinding};
    \end{scope}
    \begin{scope}[shift={(4.2,0)}]
      \draw[s, fill=olive!8] (0,0) -- (0,0.9) -- (1.35,0.9) -- (1.35,0.3) -- (2.05,0.3) -- (2.05,0.9) -- (3.4,0.9) -- (3.4,0) -- cycle;
      \draw[s, fill=magenta!18] (1.5,0.55) rectangle (1.9,1.7);
      \draw[f] (1.7,2.05) -- (1.7,1.73);
      \node[lbl] at (1.7,-0.45) {Peg-in-hole};
    \end{scope}
    \begin{scope}[shift={(0,-2.7)}]
    \draw[s] (0,0.2) .. controls (1.0,0.95) and (2.4,0.95) .. (3.4,0.2);
    \draw[s, decorate, decoration={coil, aspect=0.45, segment length=3.5pt, amplitude=3.5pt}]
      (1.25,0.70) -- (1.25,0.2);
    \draw[s] (1.0,0.2) -- (1.50,0.2);
    \draw[s, decorate, decoration={coil, aspect=0.45, segment length=3.5pt, amplitude=3.5pt}]
      (2.15,0.70) -- (2.15,0.2);
    \draw[s] (1.90,0.2) -- (2.40,0.2);
    \draw[s, fill=magenta!18] (1.7,1.05) circle (0.28);
    \draw[f] (1.7,1.7) -- (1.7,1.36) node[midway,right,lbl]{$F_n$};
    \node[lbl] at (1.7,-0.45) {Massaging};
    \end{scope}
    \begin{scope}[shift={(4.2,-2.7)}]
      \draw[s] (0,0.2) -- (3.4,0.2);
      \draw[s, fill=black!8] (1.75,0.2) rectangle (2.55,1.0);
      \draw[s, fill=magenta!18] (1.30,0.60) circle (0.20);
      \draw[f] (1.5,0.6) -- (1.73,0.6);
      \draw[f] (2.65,0.6) -- (3.25,0.6);
      \node[lbl] at (1.7,-0.45) {Non-prehensile pushing};
    \end{scope}
    \end{tikzpicture}
    \caption{{Schematic illustrations of four representative contact-rich tasks: surface grinding, peg-in-hole insertion, massaging, and non-prehensile pushing, highlighting the dominant contact configuration and force directions ($F_n$: normal contact force). }}
    \label{fig:task-images}
\end{figure}


The \textbf{definition of contact-rich tasks} varies in the literature, and it was also referred to by different terms such as physical interaction tasks~\citep{hogan2022contact}, or manipulation in contact~\citep{suomalainen2022survey}. 
According to our definition, a task is contact-rich when its successful execution requires dynamic and sustained physical contact with the environment, where motion and force are tightly coupled through contact constraints.

In this regard, for instance, obstacle avoidance is not a contact-rich task: even though contact can happen sometimes, the task can be accomplished without contact. 
Simple pick-and-place tasks are not included even though they involve contacts with objects, because these contacts are fixed (i.e., not dynamic) after the grasp. 
However, tool-use tasks where the grasped object interacts with the environment, or in-hand manipulation where the contacts between the object and the robot hand are not fixed, are considered contact-rich. 
The definition also excludes the impact-based tasks such as throwing \citep{100}, hitting \citep{48} or catching \citep{tassi2025ima} an object, in which the contact is momentary or sparse (i.e., not extended). These tasks are dynamic, but not contact-rich.

Some examples of typical contact-rich tasks are peg-in-hole insertion, surface grinding, massaging a human, multi-finger manipulation, vegetable cutting, and box pushing (Fig.~\ref{fig:task-images}). 
We advance our discussion on the contact-rich tasks in \emph{Contact-Rich Task Categories}, by grouping the tasks in the reviewed literature into four main categories, and presenting uncommon task variants that introduce novel challenges.

{The defining features of contact-rich tasks, namely dynamic, sustained contact with tightly coupled motion and force, are what make them uniquely challenging from a safety standpoint in three interrelated ways. First, the dynamics are \emph{hybrid and discontinuous}: as the system transitions between free motion, impact, sticking, and sliding, the governing equations and the set of active contact constraints switch abruptly, so a single smooth certificate (e.g., one Lyapunov function or one control barrier function) often fails to remain valid across mode boundaries, and one may instead require multiple Lyapunov functions or mode-aware certificates (see \emph{Provable safety methods})~\citep{hogan2022contact, pang2023planning, branicky2002multiple}. Second, motion and force are \emph{tightly coupled}: at contact, position-tracking and force-regulation objectives directly conflict, since commanding a position into a stiff surface produces large interaction forces, and safety therefore cannot be reasoned about in position space alone. Compliant parameterizations such as impedance and admittance control make this trade-off explicit, which is also why the learning-based methods surveyed here predominantly output pose/force/impedance references rather than raw torques.
Third, these two properties raise the \emph{relative degree} of the safety constraints: physically meaningful limits such as contact-force envelopes and friction cones depend on the contact mode and on higher-order derivatives of the configuration, which complicates the synthesis of provable safety filters and is a recurring difficulty for high-relative-degree CBFs in contact-rich settings~\citep{breeden2021high, wang2022high, wong2025contact}. 
}




\subsection{Definition of Safety in Contact-Rich Tasks}\label{sec:safety-defi}
{The definition of safety in contact-rich manipulation} spans multiple, complementary notions that arise from diverse objectives, stakeholders, and operational contexts~\cite{wabersich2023data}. Rather than a single definition, safety should be understood along several axes: physical interaction and energy exchange; stability and invariance; constraint satisfaction and reachability; risk- and data-driven guarantees; human-centered factors and standards~\cite{ISO15066}; and phase- and level-specific integration (exploration vs.\ execution; planning vs.\ control). We formalize these axes and situate them within contact-rich robotics with representative references.

\textbf{Physical interaction safety, including contact force, energy, and passivity,} forms a primary foundation for safety in contact-rich manipulation.\label{sec:safety-physical} Excessive normal or tangential forces, whether arising as high transient impacts at contact onset or as sustained interaction forces during steady contact, can damage workpieces and tooling or injure humans in collaborative settings. A practical definition of safety therefore specifies force envelopes over time, including bounds on force magnitude, rate of change, and impulse, together with region- or phase-specific limits (e.g., fragile surfaces and no-touch zones). Reviews on interaction, impedance control, and force regulation provide formal treatments of these aspects and their implications for stability~\citep{ajoudani2018progress, calanca2015review}.

Achieving force-centric safety requires (i) reliable sensing/estimation of contact forces and states, (ii) controllers that prioritize force regulation under uncertainty, and (iii) runtime mechanisms that enforce hard limits. Force/torque sensing combined with observers and multimodal perception~\cite{huang_unified_2025} supports contact-state estimation (onset, sticking/slip) and force-grounded policy inputs; this enables early guard actions (e.g., approach deceleration, alignment) to reduce peaks and avoid lateral overload. Classical hybrid {position-force} control and impedance/admittance control provide physically interpretable means to track force references while maintaining compliant interaction; variable impedance schemes adapt stiffness/damping online to keep forces within bounds amid modeling error and task variation \citep{ajoudani2018progress, calanca2015review}. In learning-integrated systems, producing high-level references (pose/force/impedance gains) rather than direct torques helps preserve stability margins while meeting force objectives \citep{ajoudani2018progress, yuan2022safe}.

Enforcement at execution time benefits from casting force bounds (and friction cones) as constraints. Model-based safety filters, such as MPC with hard/soft constraints or barrier-based action projection, can minimally modify a nominal (possibly learned) command to preserve admissible contact forces, especially near transitions (free-space to contact, contact to sliding). This projection viewpoint also supports chance/risk-aware limits to reduce rare but damaging force spikes \citep{yuan2022safe, 29, 152}. When specifications are given in natural language or task rules (e.g., “do not exceed 5 N on cable,” “avoid pressing the painted surface”), grounding them into formal force/region constraints and injecting them into the safety layer provides reliable guardrails during execution \citep{33, 100}.

Energy and passivity complement the force-first view by regulating how much energy the robot injects into the environment. Passivity indices and energy reservoirs (energy tanks) offer monitors and buffers that cap impact energy and prevent force runaway under delays, unmodeled compliance, or high-gain settings. In practice, combining passivity-aware impedance/admittance with explicit force envelopes yields robust behavior across approach, impact, and sustained-contact phases, and remains compatible with learning-based references and uncertainty-aware shields \citep{ajoudani2018progress, calanca2015review, yuan2022safe}.

\textbf{Stability and invariance, as characterized by Lyapunov and control barrier function (CBF) formulations,} provide a second perspective on safety by treating it as stability and set invariance.\label{sec:safety-invariance} Lyapunov methods certify convergence or boundedness and can imply safety when safe sets are invariant sublevel sets of a Lyapunov function~\citep{khalil2002nonlinear, kumar_safe_2025}. Control Barrier Functions (CBFs) enforce forward invariance of safety sets online by minimally modifying actions to prevent constraint violations, and have seen increasing use in contact-aware robotics and safe reinforcement learning~\citep{286, guerrier2024learning, 213, 235}. End-to-end safe RL can further embed barrier-based constraints into training or runtime filtering~\citep{287, 129}. From this perspective, safety means remaining within certified invariant sets while preserving task feasibility.

\textbf{Constraint satisfaction and reachability} provide another complementary view of safety by emphasizing the persistent satisfaction of physical constraints.\label{sec:safety-reachability} Safety is defined through state, input, and contact constraints, including joint, velocity, and torque limits, friction cones, and contact-force envelopes. Safe MPC enforces these constraints over a prediction horizon using robust tubes or uncertainty sets~\citep{hewing2020learning, 21}. Hamilton--Jacobi reachability and safety filters compute (or approximate) safe sets and veto actions that would leave them~\citep{bansal2017hamilton, wabersich2023data, 271}. Under this constraint-centric view, safety is achieved by maintaining physically meaningful bounds throughout approach, impact, sliding, and re-grasp phases.

\textbf{Risk- and data-driven safety} complements analytical formulations by quantifying safety through statistical measures of uncertainty and risk.\label{sec:safety-risk} Data-driven definitions characterize safety using risk-sensitive objectives, recovery regions, or auxiliary critics that predict and penalize unsafe outcomes. Recovery RL separates exploration from recovery to reduce unsafe excursions~\citep{91}, while safety critics and pessimistic objectives downweight risky actions under uncertainty~\citep{185, gu2024review, brunke2022safe}. Recent work further provides theoretical formulations, including safe value functions, shielding, and generalization guarantees, that connect statistical learning with formal notions of safety~\citep{165, 137}. From this perspective, safety is expressed as a probabilistic guarantee or risk bound over contact outcomes under imperfect models and noisy sensing.

\textbf{Human-centered and standard-driven safety} extends the notion of safety beyond robot performance to include human well-being and acceptability.\label{sec:safety-human} In human--robot collaboration and assistance, safety must address not only physical limits, such as contact pressure, speed-and-separation distance, and ergonomics, but also perceived safety, including predictability, comfort, and trust~\citep{ajoudani2018progress, 52}. Standards such as ISO~10218 and ISO/TS~15066 codify acceptable limits and safety procedures, shaping permissible forces, speeds, and safety-rated monitoring~\citep{280, lakshminarayanan2024robots}. Perception- and policy-level safeguards further reduce human exposure to risk during collaborative and assistive tasks~\citep{zhao2021efficientHRC, el2020towardsICRA, 57, 9}. Accordingly, safety encompasses not only protecting robots and objects but also ensuring human well-being and acceptance.

\textbf{Semantic safety enabled by vision-language(-action) models} introduces a higher level of safety reasoning by aligning task intent, environmental context, and contact constraints with robot execution.\label{sec:safety-semantic} In contact-rich settings, semantic safety requires grounding task semantics, including objects, affordances, no-touch regions, insertion axes, and force limits, together with safety rules such as force thresholds, PPE requirements, and human proximity, into actionable constraints and runtime monitors for execution~\citep{yuan2022safe}.

{It is useful to separate two notions that are easily conflated. \emph{Semantic (task-level) safety} concerns whether the model selects an appropriate, non-harmful plan, for example, refusing an unsafe instruction, identifying a fragile object or a no-touch region, and choosing a sensible approach. \emph{Physical contact safety} concerns whether the \emph{executed} interaction respects quantitative limits on force, impulse, velocity, and contact region. The two are not interchangeable: a plan can be semantically correct yet still produce an unsafe contact, and a semantically awkward plan can remain physically safe if the low-level controller bounds the interaction. The current state-of-the-art VLAs contribute to semantic safety and reasoning, including parsing intent and grounding constraints. In contrast, physical contact safety is still ensured by the downstream compliant or shielding layer (impedance control, CBF/MPC/reachability filters) rather than by the language model itself. We use ``semantic safety'' for the former and reserve force/region/impulse guarantees for the latter throughout this survey.

Modern VLM/VLA systems can (i) parse natural-language goals and safety rules, (ii) ground them in the sensed scene (objects, surfaces, grasp points) \citep{xu2025stare, 10705419, 33, 51}, and (iii) parameterize low-level compliant controllers or skill libraries with safety-aware references (pose/force/impedance) \citep{330, zhou_physvlm_2025, wei_audio-vla_2025} rather than direct torques. Typical architectures use a hierarchical split: a language–vision planner selects certified skills/options; mid-level skills expose safety-relevant parameters; low-level safety enforcers (impedance, CBF/MPC, rate limiters) guarantee contact stability and constraint satisfaction during execution \citep{wei_audio-vla_2025, cui_end--end_nodate, lu_vla-rl_2025, 376}.
{We instantiate this hierarchical split concretely in the Safety Abstraction Level section (Fig.~\ref{fig:ppe_stack}), where the plan--parameterize--enforce stack is shown with named modules and example systems, alongside a representative failure mode in which the low-level shield catches a mis-grounded language constraint.}


\textbf{A general problem formulation for contact-rich manipulation} considers tasks arising across assembly, insertion, forceful surface interaction, and human--robot collaboration.\label{sec:problem-setting} We adopt a formulation that does not commit to a single mathematical model. Tasks are described by (i) a state representation $x_t$ that includes robot kinematics and dynamics, latent contact modes, and measured or estimated wrenches; (ii) a control signal $u_t$ that may consist of low-level torques or velocities, or higher-level references such as force targets and impedance/admittance gains; and (iii) safety and feasibility specifications involving force and torque envelopes, friction and approach cones, motion limits, and human-centric constraints~\citep{ajoudani2018progress, calanca2015review, yuan2022safe}. Depending on the underlying assumptions and application requirements, the decision model may be formulated as an unconstrained Markov decision process (MDP), a partially observable Markov decision process (POMDP), a constrained Markov decision process (CMDP), or a hybrid systems/control-theoretic framework with explicit contact transitions. Across these formulations, safety is operationalized through limits on contact forces, pressures, and motion, potentially augmented by risk-aware or semantic constraints~\citep{yuan2022safe, 33, 100, 330}.

Observability and abstraction are key design choices. When direct force/torque sensing is available, controllers can track force references and tighten safety envelopes; otherwise, observers or learned estimators provide wrench/contact-state cues for policy inputs and online monitoring \citep{calanca2015review}. Action abstraction often improves safety and sim-to-real transfer: outputting pose/force targets or impedance parameters instead of raw torques yields compliant interaction and clearer hooks for runtime safety mechanisms \citep{calanca2015review, ajoudani2018progress, yuan2022safe}. Finally, semantics (VLM/VLA) supply high-level task structure and safety rules that must be grounded into formal constraints and checked during execution \citep{33, 100, 330, geng2025roboverse, gu2025robust}.

Fig.~\ref{fig:structure} provides an overview of this survey. Our review is based on the background and key definitions given at the beginning of \emph{Robotic Contact-Rich Tasks and Definition of Safety in Contact-Rich Tasks}. The core of our review is organizing related work from multiple perspectives, covering task characteristics, sensing and policy modalities, data acquisition strategies, simulation environments and benchmarks. Our results are presented as the identified key challenges and new perspectives.
\begin{figure}
    \centering
    \begin{overpic}[width=0.92\linewidth]{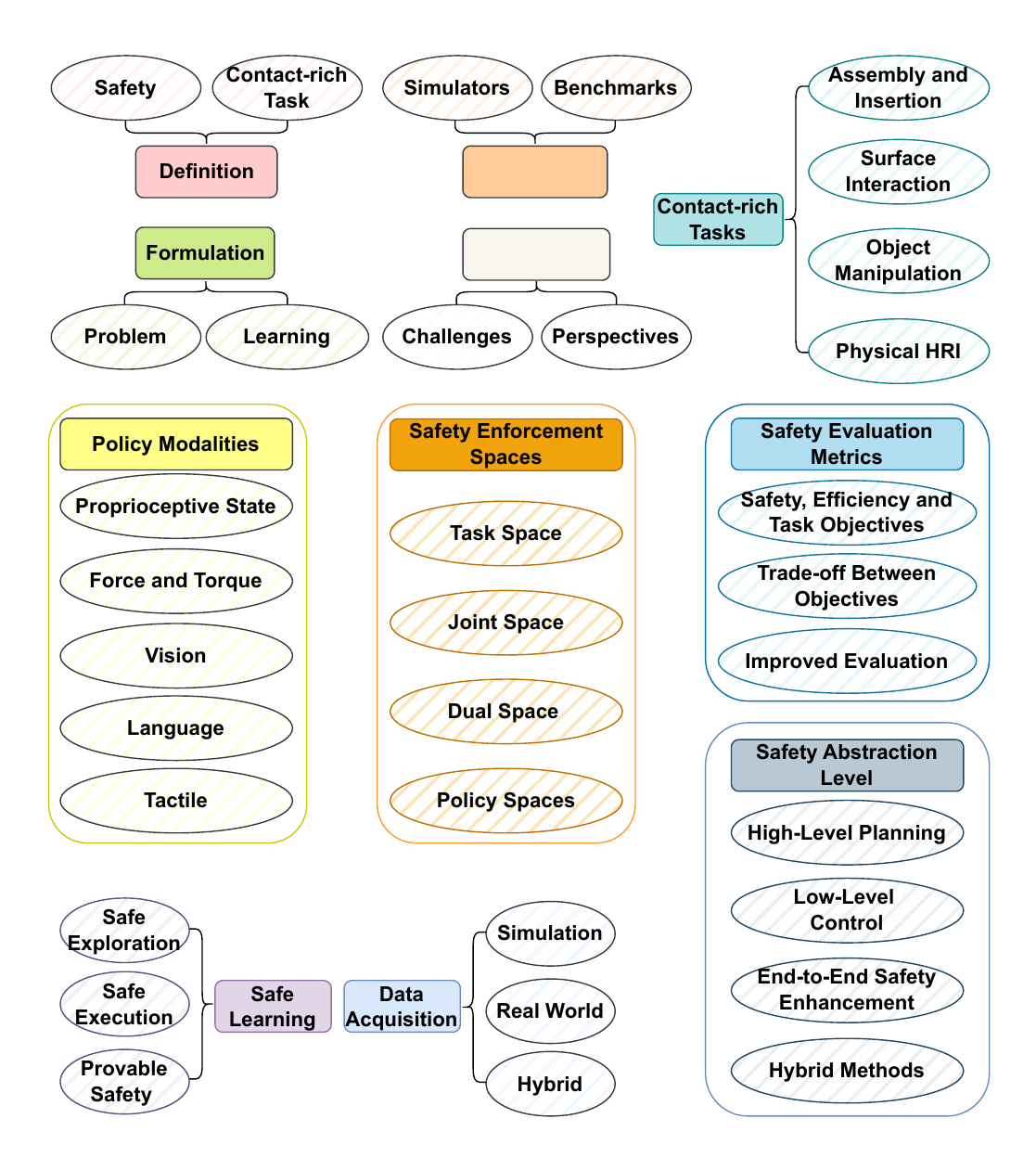}
        \put(40,84.5){\scriptsize  Simulation}
        \put(40,77.5){\scriptsize  Challenges}
    \end{overpic}
    \caption{Structure of this survey. We introduce the background and definitions in the beginning of, and categorize existing methods from different perspectives, including task characteristics, sensing and policy modalities, data acquisition strategies, simulation environments and benchmarks, safety evaluation metrics, safety abstraction levels, and safety enforcement spaces. Our results consist of the discussion of open challenges and new perspectives.}
    \label{fig:structure}
\end{figure}

\textbf{Learning paradigms for safe contact-rich manipulation} can be broadly categorized into reinforcement learning, imitation learning, learning-enhanced model-based control, foundation-model-based methods, and hybrid integration, each interfacing with safety through different mechanisms.\label{sec:learning-paradigm}

\textbf{Reinforcement learning} learns policies for variable-impedance forceful skills and insertion tasks, often using action spaces that output force targets or stiffness and damping parameters to preserve compliant interaction. Safety is enhanced through conservative objectives, recovery strategies, and shielded execution mechanisms, such as barrier-function, reachability-based, and MPC-based safety layers~\citep{zhang2024srl, roveda2020model, stan2020reinforcement, zhang2025passivity, 348, 320}.

\textbf{Imitation learning} acquires initial policies for contact sequencing and force modulation (e.g., approach, touch, and sliding) from demonstrations. These policies are commonly refined through reinforcement learning or supervised updates, while hybrid trajectory-force learning has become prevalent in insertion and assembly tasks~\citep{246, 27, luu2025manifeel}. In this survey, we focus on methods with a substantial online learning component and refer readers to recent surveys for purely imitation-learning approaches~\citep{Tsuji2025asu}. Although VLA policies are typically trained through imitation learning, we include them under the criterion defined in the Robotic Contact-Rich Tasks section, because they incorporate explicit safety mechanisms and/or online refinement rather than simply replaying demonstrations.

\textbf{Learning-enhanced model-based and control-theoretic methods} combine learning with established control frameworks, including impedance and admittance regulation, robust and adaptive force control, MPC with contact constraints, and barrier- or reachability-based safety filters. In these approaches, learning augments classical controllers by identifying system dynamics, adapting controller gains, or shaping feasible sets, thereby improving adaptability while retaining transparent runtime safety enforcement~\citep{242, 235, 318, pang2023planning, xu_look_2021}.

\textbf{Foundation-model-based methods} leverage large-scale vision-language (and action) pretraining to interpret open-ended task descriptions, infer affordances, and generate safety-aware plans before execution. In contact-rich settings, these methods are commonly organized into a plan--parameterize--enforce pipeline: (i) a VLM/VLA interprets natural-language goals and scene context to generate task sketches and safety rules (e.g., force thresholds, no-touch regions, and speed limits); (ii) the resulting semantic plan is translated into formal constraints and parameterized skill references (pose, force, or impedance) rather than raw torque commands to preserve compliant interaction~\citep{xu2025stare}; and (iii) certified low-level safety layers, such as MPC or CBF filters and rate limiters, execute the policy while enforcing bounds on contact forces and approach geometry~\citep{33, 100, 330}. We discuss these safety guardrails in greater detail in the Safety Abstraction Level section, together with representative systems, an architectural overview, and a worked failure case.

\textbf{Hybrid integration} combines complementary strengths across planning, learning, and formal safety. High-level planning and skill graphs, potentially guided by VLMs or VLAs, provide semantic structure, while learned policies generate pose, force, or impedance references and certified safety layers project actions into admissible sets. Semantic reasoning modules further align high-level task intent and safety rules with the certified execution stack~\citep{geng2025roboverse, gu2025robust, 330, 10705419}. Performance is commonly evaluated using standardized simulation suites and benchmarks that report safety violations, near misses, and generalization performance~\citep{towers2024gymnasium, raffin2021stable, 152}.

In this section, we categorize existing \textbf{safe learning approaches} for contact-rich robot tasks based on the phase: safe exploration (learning before contact) and safe execution (learning during contact) and provable safety methods, and discuss representative methods in each category.
This categorization helps clarify the distinct challenges and strategies associated with each phase, as well as their implications for safety in contact-rich manipulation.

\textbf{Safe exploration, corresponding to learning before contact,} is the most safety-critical phase of contact-rich robot learning because unsafe exploration can lead to collisions, excessive contact forces, and hardware damage before a reliable policy is obtained. This section reviews representative methods for safe exploration and discusses how they improve learning efficiency while minimizing physical risk prior to task execution.

Safe human-robot cooperation (HRC) is a major application branch within the broader class of contact-rich tasks, and it has traditionally been approached through path planning algorithms such as rapidly-exploring random tree star (RRT$^*$)~\cite{chi2017risk} for dynamic human-robot systems, or artificial potential field (APF)~\cite{rodriguez2023advanced} and virtual fixtures~\cite{bowyer2013active} for real-time adjustment of the robot's trajectory. 
However, these methods often yield slow trajectory computation and limited generalization, since they require task-specific tuning, assume simplified environments, and adapt poorly to dynamic or unstructured settings~\cite{bowyer2013active}.

Recent approaches enforce safety constraints via control barrier functions (CBFs), where quadratic programs compute minimal trajectory deviations to maintain safety~\cite{hsu2023safety, ferraguti2020safety, dai2023safe}. Extensions to uncertain CBFs (U-CBFs) handle environmental uncertainty~\cite{das2024robust, li2023robust}, though worst-case assumptions often yield conservative designs. Stochastic frameworks using chance constraints offer less conservative alternatives~\cite{wang2024flow, du2011probabilistic}, but remain computationally intractable for most HRC applications due to non-convex feasible regions~\cite{ahmed2008solving}. Complementary work examines risk factors in HRC~\cite{berx2022identification, inam2018risk, liu2020dynamic, majumdar2020should}. Although these approaches can better characterize tolerable risk bounds for different applications, risk-tunable CBFs can further reduce conservatism~\cite{11023035}. 

Ensuring system safety while maintaining performance requires sophisticated approaches that balance computational efficiency, safety guarantees, and adaptability. Below, we present four prominent categories of methods that address these challenges through different theoretical frameworks and computational strategies.


Across these reachability-based methods, a common theme is to shift most of the computational burden into the offline construction of trajectory-parameterized reachable sets based on a reasonably accurate model. Reachability-based Trajectory Safeguard (RTS)~\cite{271} is an offline--online safety layer for continuous-control RL that precomputes trajectory-parameterized forward reachable sets and, at runtime, projects the RL agent's chosen plan parameters to a nearest safe plan while preserving learning efficiency. A reachability-based manipulator planner~\cite{Holmes-RSS-20} builds joint-space, trajectory-parameterized reachable sets offline and then uses them to constrain receding-horizon optimization. Another paper~\cite{kousik2020bridging} introduces a receding-horizon planner for mobile robots that performs offline reachability analysis over a continuum of trajectory parameters and then selects safe parameters online. Those approaches rely on a reasonably accurate system model and can be less adaptive when dynamics or obstacles change significantly during operation.

Model predictive shielding (MPS)~\cite{banerjee2024dynamic} provides a model-predictive shield around any RL policy: it forecasts violations and replaces unsafe actions with safe, goal-directed recoveries, enabling zero-violation training and deployment. CBF-Guided MPC~\cite{213} embeds barrier constraints in MPC and tunes them online for proactive, explainable hard-safety during tracking/avoidance at the cost of online optimization and careful model or CBF design.
These MPC-based methods trade offline precomputation for online optimization to improve task progress with interpretable constraints. Planner-augmented approaches such as MoPA-RL \cite{99} can be viewed as a broader shield that uses classical motion planning (and reactive shaping) to intervene and generate collision-free long-horizon actions around a learned policy, improving exploration and execution safety without CBF-style hard guarantees.

RTS-style reachability safeguards shift burden offline, enabling fast runtime projection with hard safety so long as the precomputed reachable sets remain valid; their main limitation is reduced adaptivity under model mismatch or rapidly changing contact dynamics. In contrast, MPS and CBF-guided MPC place more computation online: they can respond to changing constraints and goals within a receding horizon, but require repeated optimization and careful dynamics/constraint modeling, with feasibility and real-time performance becoming the bottleneck in high-dimensional HRC. Practically, reachability filters suit fixed platforms with stable dynamics and tight real-time budgets, whereas MPC/CBF variants suit environments where constraints and intent change quickly and online compute is available.

Implicit Safe Set Algorithm~\cite{zhao2024implicit}.  A model-free safe control method that queries a black-box dynamics (e.g., digital twin) to synthesize a safety index (barrier certificate) and a corresponding safe control law, guaranteeing forward invariance and finite-time convergence. 
Data-Driven Permissible Safe Control with Barrier Certificates~\cite{121} learns unknown stochastic dynamics via Gaussian Processes and constructs piecewise stochastic barrier functions to compute a maximal set of permissible strategies that satisfy a target probabilistic safety level; as data increases, the permissible set grows toward the known-model benchmark while maintaining safety. Together with safe exploration methods \cite{sootla2022saute, 275, lei2024langevin}, these approaches illustrate how pre-deployment safe learning can constrain exploration to certified-safe behaviors before execution.

\begin{figure}[t]
    \centering
    \includegraphics[width=1\linewidth]{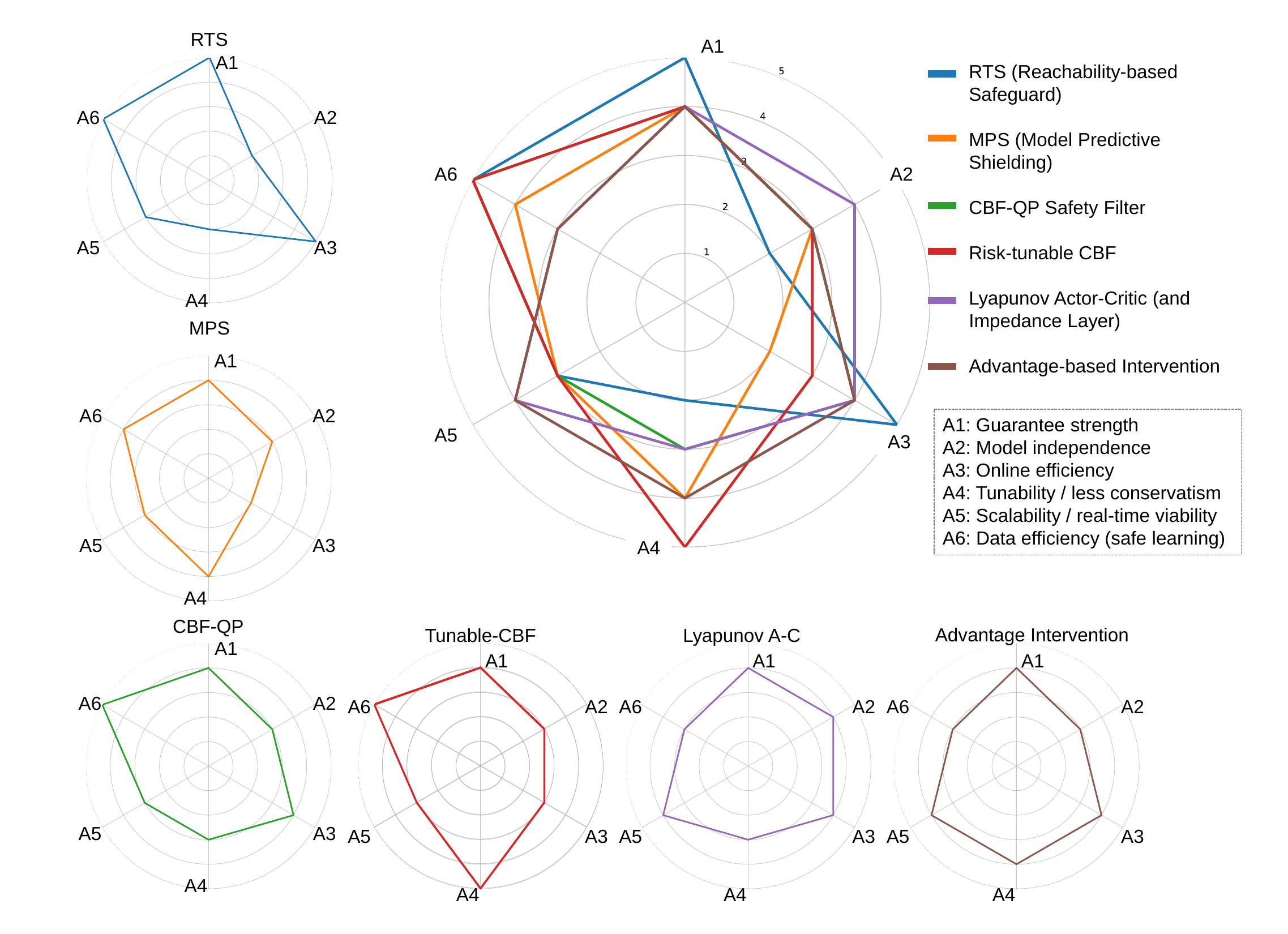}
    \caption{
    Hexagonal ratings across six comparison axes (higher values indicate stronger relative performance on a 1--5 scale).
    (A1) Guarantee strength: 1 = empirical/heuristic only, 5 = hard safety (invariance/reachability-style).
    (A2) Model independence: 1 = requires an accurate known model, 5 = model-free / black-box / learned abstraction.
    (A3) Online efficiency: 1 = heavy online optimization, 5 = lightweight runtime (most computation offline).
    (A4) Conservatism / tunability: 1 = highly conservative with few tuning knobs, 5 = explicitly risk-tunable and typically less conservative in practice.
    (A5) Scalability / real-time viability: 1 = struggles in high-dimensional contact-rich or multi-agent settings, 5 = scales well and runs in real time.
    (A6) Data efficiency (safe learning): 1 = data-hungry, 5 = requires little data to ensure safety or supports efficient safe data collection.
    {These ratings are an explicit, rubric-based \emph{relative} synthesis, not benchmark measurements: the 1--5 anchors for each axis are defined in Table~\ref{tab:rubric}, the per-method rationale and supporting evidence are given in Table~\ref{tab:rationale}, and each score was assigned independently by two of the authors using that rubric and reconciled by discussion.}
    }
    \label{fig:6dim_hex_ratings}
\end{figure}

Advantage-Based Intervention~\cite{wagener2021safe} achieves safety by online interventions and surrogate MDP training (requires a safe backup policy/advantage estimates), whereas AlwaysSafe~\cite{simao2021alwayssafe} achieves safety by a model-based abstraction of safety dynamics (requires a correct abstraction); both enable safe data collection for learning.

Reachability filters policy actions via offline reachable sets for real-time, hard-safety execution; MPC/CBF enforces safety online via predictive checks and constraints with interpretable but optimization-heavy control; safety-set constructs invariant or permissible safe regions model-free or GP-based for certified operation; and safe exploration guarantees no violations during learning via interventions or abstract safety models, trading flexibility and sample efficiency against model/backup policy assumptions.

Here we group learning-based methods by what they learn and how they ensure safety. Lyapunov actor–critic~\cite{16} learns policy weights under an explicit Lyapunov constraint, and executes through an adaptive impedance layer for compliant, stability-guaranteed motion. Safety-Aware Unsupervised Skill Discovery~\cite{22} learns a skill policy and a {safety critic} that estimates failure probability, yielding intrinsically safe skills for downstream tasks. Variable-Impedance Skill Learning~\cite{23} learns a latent skill prior/decoder, a latent policy, and impedance as part of the action, achieving safety via compliant interaction and bounded contact forces. Impedance-learning adaptive control for HRI~\cite{9531394} learns impedance gains, feedforward interaction force, dynamics parameters, and robust Sliding Mode Control (SMC) gain, with Lyapunov proofs of bounded tracking. Impedance iterative-learning SMC~\cite{5} updates impedance and SMC gains across trials, combining compliance and robustness guarantees for non-repetitive tasks. Model-accelerated maximum-entropy RL for high-precision assembly~\cite{98} learns a stochastic policy executed through an impedance controller and leverages a GP dynamics model for sample-efficient training, achieving practical safety via compliant interaction.

 Learning-based approaches directly optimize policies, skills and impedance from data, offering greater adaptability to unmodeled dynamics and complex contact phenomena with less hand-crafted modeling, most advantageous when rich demonstrations or interactive rollouts are available and the task/environment is hard to model yet demands fast, compliant behavior under distribution shift.

\begin{table*}[t]
    \centering
    {
    \footnotesize
    \renewcommand{\arraystretch}{1.25}
    \caption{{Scoring rubric for the hexagonal ratings of Fig.~\ref{fig:6dim_hex_ratings}. For each axis we give the anchors for scores~1, 3, and~5 on the relative 1--5 scale; scores of~2 and~4 denote intermediate cases that interpolate between adjacent anchors. The ratings are an ordinal, relative synthesis across the cited studies, not absolute benchmark measurements.}}
    \label{tab:rubric}
    \begin{tabularx}{\textwidth}{@{}l >{\raggedright\arraybackslash}X >{\raggedright\arraybackslash}X >{\raggedright\arraybackslash}X@{}}
    \toprule
    \textbf{Axis} & \textbf{Score 1} & \textbf{Score 3} & \textbf{Score 5}\\
    \midrule
    A1 Guarantee strength & Empirical / heuristic only & Stability- or probability-based guarantee & Hard safety (set invariance / reachability)\\
    A2 Model independence & Requires an accurate known model & Approximate or partially identified model & Model-free / black-box / learned abstraction\\
    A3 Online efficiency & Heavy online optimization & Moderate per-step optimization (e.g., a QP) & Lightweight runtime (most computation offline)\\
    A4 Conservatism / tunability & Highly conservative, few tuning knobs & Some tunable safety margins & Explicitly risk-tunable, less conservative in practice\\
    A5 Scalability / real-time & Struggles in high-dimensional contact-rich or multi-agent settings & Moderate (limited horizon or contact count) & Scales well and runs in real time\\
    A6 Data efficiency & Data-hungry & Moderate data requirement & Little data needed, or supports efficient safe data collection\\
    \bottomrule
    \end{tabularx}
    }
\end{table*}

\begin{table*}[t]
    \centering
    {
    \footnotesize
    \renewcommand{\arraystretch}{1.25}
    \caption{{Per-method rationale for the hexagonal profiles in Fig.~\ref{fig:6dim_hex_ratings}, scored with the rubric of Table~\ref{tab:rubric}. We give the qualitative justification and supporting references for each method family; the plotted 1--5 values themselves are shown in Fig.~\ref{fig:6dim_hex_ratings}. Two of the authors assigned scores independently using the rubric and reconciled disagreements by discussion.}}
    \label{tab:rationale}
    \begin{tabularx}{\textwidth}{@{}l >{\raggedright\arraybackslash}X@{}}
    \toprule
    \textbf{Method family} & \textbf{Rationale for its A1--A6 profile (with supporting references)}\\
    \midrule
    RTS (reachability-based safeguard) & Strong guarantee (A1: hard reachability invariance) and online efficiency (A3: reachable sets precomputed offline, lightweight online projection), and data-efficient (A6: model-based safety preserves RL learning); lower model independence (A2: needs a reasonably accurate model) and tunability (A4: few online knobs, conservative under mismatch) \citep{271, Holmes-RSS-20, kousik2020bridging}.\\
    MPS (model predictive shielding) & Strong guarantee (A1: forecasts and substitutes unsafe actions for zero violation) and adaptivity within the horizon (A4); lower online efficiency (A3: receding-horizon optimization each step) and model independence (A2: needs a predictive model) \citep{banerjee2024dynamic, bastani2021safe}.\\
    CBF-QP safety filter & Strong guarantee (A1: forward invariance when assumptions hold) and good online efficiency (A3: a cheap per-step QP); can be conservative under uncertainty (A4 low for worst-case U-CBFs) and depends on an accurate model (A2 low) \citep{hsu2023safety, ferraguti2020safety, dai2023safe}.\\
    Risk-tunable CBF & Like the CBF-QP filter but with explicit risk tuning (A4 high, less conservative) at a probabilistic rather than hard guarantee (A1 moderate); still an online QP (A3 moderate) \citep{11023035}.\\
    Lyapunov actor--critic (+ impedance) & Strong model independence (A2: largely model-free policy with an impedance layer) and lightweight execution after training (A3); guarantee is stability-based rather than hard invariance (A1 moderate) and effectiveness depends on training coverage (A6 lower) \citep{16, 23, han2020actor}.\\
    Advantage-based intervention & Designed for safe exploration: supports efficient safe data collection (A6) and is model-free (A2 high); guarantee depends on a safe backup policy (A1 moderate) and conservatism is set by the intervention frequency (A4 moderate) \citep{wagener2021safe, 91}.\\
    \bottomrule
    \end{tabularx}
    }
\end{table*}

Since the literature lacks a unified benchmark that evaluates different safety paradigms on the same contact-rich task under matched conditions, the comparisons in this section are necessarily synthesis-based: we distill the reported assumptions, guarantees, and practical constraints from the above studies into a common set of criteria. To provide a unified lens for analyzing trade-offs across method families, we compare them using six recurring axes:
(A1) guarantee strength/type (hard invariance/reachability vs.\ probabilistic vs.\ empirical),
(A2) model reliance (known dynamics vs.\ approximate/black-box vs.\ data-driven),
(A3) compute placement (offline precomputation vs.\ online optimization/filtering),
(A4) conservatism and risk tuning (worst-case bounds vs.\ chance levels vs.\ tunable safety margins),
(A5) scalability and real-time viability (state dimension, horizon length, contacts/multi-agent),
and (A6) data/sample efficiency (how safety constraints affect exploration and learning speed).
In Fig.~\ref{fig:6dim_hex_ratings}, we summarize these axes using hexagonal ratings for representative methods.
{To complement the qualitative hexagonal view with an explicit, inspectable basis for these scores, Table~\ref{tab:method_comparison} compiles the underlying criteria for the same representative method families, guarantee type, model assumptions, compute placement and cost, conservatism/tunability, and data efficiency, so that the figure and the table are mutually consistent. The table columns map to axes A1 (guarantee strength/type), A2 (model reliance), A3 (compute placement), A4 (conservatism/tunability), and A6 (data/sample efficiency); axis A5 (scalability/real-time viability) is discussed qualitatively in the synthesis paragraph below and reflected in Fig.~\ref{fig:6dim_hex_ratings}.}

Across the six axes (A1--A6), the literature reveals a recurring trade-off between guarantee strength and adaptability/compute. Classical planning and reactive shaping are intuitive but often provide weaker or indirect guarantees and require task-specific tuning, limiting scalability in dynamic HRC. CBF-style methods offer explicit, interpretable safety enforcement with strong guarantees when modeling assumptions hold, but can become conservative under uncertainty (e.g., worst-case U-CBFs) or computationally heavy when using less conservative probabilistic constraints. Learning-based compliant skills improve adaptability to complex contact phenomena and can achieve practical safety via compliance and stability arguments, but their effective safety depends on training coverage and may trade formal hardness for data requirements. Reachability-based safeguards deliver the strongest hard-safety behavior with predictable runtime by shifting computation offline, yet rely on accurate models and can be less adaptive under mismatch or rapidly changing environments. MPC/shielding approaches sit between these extremes, improving adaptivity through online prediction and constraint handling but incurring repeated optimization costs and feasibility concerns in high-dimensional contact-rich settings. Safe-set/certificate construction and intervention/abstraction-based safe exploration further enable pre-deployment or training-time safety, but their conservatism and scalability are governed by certificate fidelity, backup/abstraction correctness, and how often safety mechanisms override exploration.

\newcolumntype{Y}{>{\RaggedRight\arraybackslash}X}

\begin{sidewaystable}[t]
    \centering
    {
    \footnotesize
    \renewcommand{\arraystretch}{1.25}
    \caption{{Structured comparison of representative safe-learning method families for contact-rich tasks.}}
    \label{tab:method_comparison}

    \begin{tabularx}{\textwidth}{Y Y Y Y Y Y Y}
    \toprule
    \textbf{Method family} &
    \textbf{Guarantee type (A1)} &
    \textbf{Model assumptions (A2)} &
    \textbf{Compute placement \& cost (A3)} &
    \textbf{Conservatism / tunability (A4)} &
    \textbf{Data efficiency (A6)} &
    \textbf{Refs.} \\
    \midrule

    RTS (reachability-based safeguard) &
    Hard (forward-reachability invariance) while precomputed sets remain valid &
    Requires a reasonably accurate model for offline reachable-set construction &
    Heavy offline precomputation; lightweight online projection &
    Conservative under model mismatch; few online knobs &
    High: preserves RL learning efficiency; model-based safety needs little data &
    \cite{271}, \cite{Holmes-RSS-20}, \cite{kousik2020bridging} \\

    MPS (model predictive shielding) &
    Hard online: forecasts violations and substitutes safe goal-directed recoveries &
    Requires a predictive (forward) model &
    Online receding-horizon optimization at each step &
    Adapts to changing goals/constraints within the horizon &
    Model-based; moderate &
    \cite{banerjee2024dynamic} \\

    CBF-QP safety filter &
    Hard (forward invariance) when modeling assumptions hold &
    Needs an accurate model; uncertainty handled via robust/adaptive CBFs &
    Lightweight per-step quadratic program &
    Can become conservative under uncertainty &
    Analytic, not data-driven &
    \cite{hsu2023safety, ferraguti2020safety, dai2023safe} \\

    Risk-tunable CBF &
    Probabilistic / explicitly tunable risk bound &
    Model-based CBF with a risk parameterization &
    Online quadratic program &
    Explicitly risk-tunable, less conservative &
    Moderate &
    \cite{11023035} \\

    Lyapunov actor--critic (+ impedance layer) &
    Stability-based (explicit Lyapunov constraint) &
    Largely model-free / data-driven policy &
    Lightweight online execution after training &
    Depends on training coverage &
    Data-dependent &
    \cite{16} \\

    Advantage-based intervention &
    Safe data collection via online intervention &
    Model-free; requires a safe backup policy &
    Lightweight online intervention check &
    Conservatism set by intervention frequency &
    Enables safe data collection &
    \cite{wagener2021safe} \\

    \bottomrule
    \end{tabularx}
    }
\end{sidewaystable}


\textbf{Safe execution, which focuses on maintaining safety after contact has been established,} is crucial in contact-rich robotic tasks because robots must interact not only with complex and uncertain environments but often also in close proximity to humans~\citep{ajoudani2018progress}. During these interactions, unsafe behaviors can damage the robot, the manipulated objects, or even endanger human safety. Ensuring safe execution therefore requires that robot actions consistently satisfy physical constraints~\cite{141}, avoid hazardous situations, and maintain robust performance despite uncertainties and disturbances. These requirements become particularly critical in real-world deployments, where unpredictable events and unmodeled dynamics are inevitable. Consequently, developing methods that provide reliable safety guarantees during execution is essential for the trustworthy deployment of robots in contact-rich environments.

A widely adopted and effective approach to enhance safe execution in contact-rich robotic tasks is the use of compliant controllers~\cite{317, hogan1985impedance, 9531394, 23, samadikhoshkho2025review}. These controllers enable robots to interact safely with their environment and with humans by allowing controlled flexibility and adaptability in response to external forces. Among the most common compliant control strategies are impedance control and admittance control. Impedance control regulates the dynamic relationship between the robot's motion (position or velocity) and the forces it experiences, effectively making the robot behave like a virtual mass-spring-damper system~\cite{hogan1985impedance}. This allows the robot to absorb impacts and maintain stable contact with uncertain or moving surfaces. Admittance control, on the other hand, measures external forces and commands the robot to move accordingly, making it suitable for tasks where the environment imposes significant forces on the robot~\cite{232, 205}. Both approaches are fundamental in ensuring that robots can safely and robustly perform manipulation tasks involving physical contact, reducing the risk of damage or injury during interaction.

One key challenge in ensuring safe execution is properly tuning controller parameters (such as stiffness, damping, and inertia) to suit different task scenarios and contact conditions~\cite{calanca2015review}. Incorrect parameter settings can lead to unsafe behaviors, such as excessive contact forces, instability, or failure to accomplish the task. Traditionally, these parameters are manually tuned based on expert knowledge and trial and error, which is time-consuming and may not generalize well across diverse tasks or environments. Learning-based methods offer a promising solution by automatically adapting controller parameters through data-driven optimization, enabling robots to achieve both safety and high performance in varying and uncertain contact-rich scenarios~\cite{cruz2024learning, 10705419}.

A major research direction in safe execution for contact-rich tasks is the use of RL to automatically tune the parameters of compliant controllers~\cite{elguea2023review}, such as impedance or admittance control. In these approaches, RL algorithms are employed to learn optimal values for parameters such as stiffness~\cite{zhang2024srl}, damping, and inertia, enabling the robot to adapt its compliance to varying task requirements and contact conditions~\citep{martinez2015safe}. This data-driven adaptation helps maintain safety by minimizing excessive contact forces, preventing instability, and improving task performance.

Recent research has demonstrated the effectiveness of RL-based parameter learning in a variety of contact-rich scenarios. For example, RL has been used to adjust impedance parameters online to ensure safe and robust assembly~\cite{stan2020reinforcement, 192, 49, 174}, insertion~\cite{55, 279, 344}, and surface interaction tasks~\cite{61}. Some works incorporate safety constraints directly into the RL framework, such as limiting the maximum allowable force or ensuring stability margins during learning~\cite{287, 66, zhang2024srl}. Others combine RL with model-based safety filters~\cite{tang2024learning} or control barrier functions~\cite{zhang_enhancing_nodate} to guarantee constraint satisfaction throughout the learning and execution process~\cite{zhang2024srl}.

These methods offer several advantages: they reduce the need for manual parameter tuning, allow for adaptation to unmodeled dynamics or changes in the environment, and can explicitly balance safety and performance objectives. However, challenges remain in ensuring sample efficiency, generalization across tasks, and providing formal safety guarantees during both training and deployment. Ongoing research continues to address these issues by integrating uncertainty estimation, risk-sensitive objectives, and hybrid learning-control architectures.

{Large vision-language(-action) ``robot foundation'' models also contribute to safe execution by converting unstructured scene and instruction inputs into structured, verifiable plans that parameterize compliant low-level skills. Because these mechanisms recur across the survey, we consolidate their detailed treatment in the plan--parameterize--enforce architecture, concrete example systems, and a worked failure mode in the Safety Abstraction Level section, rather than repeating it here.}


\textbf{Provable safety methods} provide the theoretical foundation for understanding and certifying safe learning in contact-rich robotic tasks.\label{sec:provable} Long before learning-based approaches became prevalent, robotic safety was extensively studied within classical control through mathematically rigorous frameworks for stability and safety analysis~\citep{hogan1985impedance, khalil2002nonlinear, bansal2017hamilton, hogan2022contact}. These include Lyapunov-based methods, control barrier functions (CBFs), contraction metrics, reachability analysis, and invariant-set theory, which offer formal guarantees of stability, safety, and constraint satisfaction~\citep{blanchini1999set, blanchini2008set, slotine1991applied, kumar_safe_2025}. Although originally developed for model-based control, many of these techniques have increasingly been integrated into modern learning-based frameworks~\citep{wabersich2023data, 35, 75, 14, 57, 80}. The following subsections review these foundational approaches and discuss their role in enabling provably safe learning for contact-rich robotic manipulation.

\textbf{Lyapunov-based methods} build on the concept of a Lyapunov function, a scalar energy-like function $V(\boldsymbol{x})$ that decreases along trajectories of a dynamical system. Formally, consider a system $\dot{\boldsymbol{x}}={f}(\boldsymbol{x}, \boldsymbol{u})$ with state $\boldsymbol{x} \in \mathcal{X}$ and control input $\boldsymbol{u} \in \mathcal{U}$. A continuously differentiable function $V: \mathbb{R}^{n} \to \mathbb{R}_{\ge 0}$ is a Lyapunov function for a closed-loop system if it is positive definite ($V(\boldsymbol{0}) =0$, $V(\boldsymbol{x}) > 0$ for $\boldsymbol{x} \neq \boldsymbol{0}$) and its time derivative is negative semi-definite ($\dot{V}(\boldsymbol{x}) \le 0 \, \forall x$).  For safety, Lyapunov analysis is often used to certify invariant sets. If we can identify a set of safe states $\mathcal{S} (e.g. \, \mathcal{S} = \{\boldsymbol{x}\mid V(\boldsymbol{x}) \le c\})$ and a Lyapunov-like function $V(\boldsymbol{x})$ that decreases or stays constant on $\mathcal{S}$, then any trajectory starting in $\mathcal{S}$ will remain in $\mathcal{S}$ for all time, making $\mathcal{S}$ a forward-invariant safe set \citep{khalil2002nonlinear}.

Lyapunov-based methods are well-suited to scenarios where the main objective is stability, for example, regulating contact force to a desired value. Nearly all classical robot controllers, e.g., PD controllers or compliance controllers, are derived with an implicit Lyapunov function to prove stability \citep{16, 2, 32}. Passivity-based control is another Lyapunov-related approach. By designing the closed-loop system to be passive, one can guarantee stable interactions, which means the robot can safely interact with the environment because it will not generate energy that could cause instability \citep{van2014port, forni2018differential, hill2022dissipativity, 35, 75}.

Despite their broad use, a major challenge for Lyapunov methods is finding a suitable Lyapunov function for complex nonlinear systems \citep{khalil2002nonlinear}. For high-dimensional robots with complex contact dynamics, hand-designing a $V(\boldsymbol{x})$ that decreases in all conditions, especially across contact/non-contact modes, is difficult. In hybrid systems, standard Lyapunov theory does not directly apply and one may need multiple Lyapunov functions \citep{branicky2002multiple, giesl2015review, manchester2017control}. Another limitation is that classical Lyapunov methods are primarily designed to certify stability or asymptotic convergence of the closed-loop system. Unless the Lyapunov sublevel sets are explicitly constructed to be forward invariant and contained in a prescribed safe set, Lyapunov stability alone does not guarantee that state or input constraints are respected along the entire transient \citep{romdlony2016stabilization, li2023survey}. In particular, a Lyapunov-stable controller may drive a robot back to its desired equilibrium after a disturbance but still allow temporary violations of safe force or workspace limits \citep{manjunath2021safe, wu2015safety}. Therefore, Lyapunov-based designs are well-suited for ensuring long-term stability or boundedness, but typically need to be augmented with barrier-type conditions to enforce state constraints and safety \citep{286, blanchini1999set}.

\textbf{Control barrier functions (CBFs)} provide a formal, Lyapunov-like tool specifically for enforcing safety constraints. While Lyapunov functions certify the stability of an equilibrium, barrier functions certify forward invariance of a safe set, which proves the system will never enter certain unsafe regions \citep{romdlony2016stabilization, ames2016control, 286}. The typical setup is as follows: Define a continuously differentiable scalar boundary function $h(\boldsymbol{x})$ such that the safe set $\mathcal{C}$ is $\{\boldsymbol{x} \mid h(\boldsymbol{x}) \ge 0\}$ and the unsafe set is $\{\boldsymbol{x} \mid h(\boldsymbol{x}) < 0\}$. As an example, the boundary $h(\boldsymbol{x})=0$ can describe the contact force reaching a limit. A valid CBF satisfies a condition that ensures $h(\boldsymbol{x})$ cannot decrease into the negative unsafe region. One common form: There exists an extended class-$\mathcal{K}$ function $\alpha(\cdot)$ such that for all states on the boundary of $\mathcal{C}$, we can choose control inputs $\boldsymbol{u}$ to yield $\dot{h}(\boldsymbol{x}, \boldsymbol{u}) \ge -\alpha(h(\boldsymbol{x}))$. If this holds, then whenever $h(\boldsymbol{x})$ is zero or positive, its derivative cannot drive $h(\boldsymbol{x})$ across the zero boundary into the unsafe region. 
Geometrically, $h(\boldsymbol{\boldsymbol{x}})$ can be seen as a "barrier" that goes to zero at the safety boundary and typically to negative values outside the safe set. 
The CBF condition forces the controlled dynamics to hover at or increase $h(\boldsymbol{x})$ at the boundary, never allowing a drop below zero \citep{wu2015safety}.

CBFs are especially powerful for real-time enforcement of safety constraints. Unlike open-loop potential field methods \citep{batinovic2022path, pan2021improved}, CBFs offer a guarantee of safety because the CBF condition is a hard constraint on the control input. 

Despite their strengths, CBFs come with several challenges. First, they require a reasonably accurate model of the system and environment. Model uncertainty or external disturbances can break the guarantee. There is active research on robust CBFs and adaptive CBFs to account for uncertainty, but guaranteeing safety under uncertainty often leads to conservative controllers \citep{pati2023robust, buch2021robust, wang2025safe}. Second, there can be conflicts between multiple CBFs or between a CBF and control goals. It is possible that no single controller can simultaneously satisfy all constraints and achieve the desired motion. In such cases, typically one should relax constraints \citep{isaly2024feasibility, breeden2023compositions}. Third, the design of CBFs becomes significantly more involved for safety constraints with high relative degree, such as contact and force constraints that depend on higher-order derivatives of the state, which are ubiquitous in contact-rich robotic tasks \citep{breeden2021high, wang2022high, wong2025contact}.


\textbf{Contraction metrics} offer a fundamentally different viewpoint than the  Lyapunov-based and CBF methods that provide scalar certificates for stability or safety with respect to particular equilibria or safe sets. Rather than analyzing the behavior of trajectories relative to a fixed point or a boundary, contraction theory studies the evolution of the differential distance between arbitrary trajectories of a dynamical system. A system is said to be contracting if infinitesimal displacements between trajectories converge exponentially to zero under an appropriate Riemannian metric. When such a contraction metric exists, all trajectories within a contraction region converge exponentially toward one another, independent of their initial conditions. This yields a strong form of incremental stability: not only do trajectories converge to a specific equilibrium, but they converge toward a unique flow or invariant trajectory structure of the system. The contraction metric thus serves as the formal certificate ensuring this incremental convergence and generalizes classical Lyapunov analysis to trajectory-to-trajectory stability \citep{liu2025certified, lohmiller1998contraction, slotine2003modular, manchester2017control}.

A metric is typically given by a positive definite matrix function $\boldsymbol{M}(\boldsymbol{x})$ defining a weighted norm $|\delta \boldsymbol{x}|_{\boldsymbol{M}}^2 = \delta \boldsymbol{x}^{\top} \boldsymbol{M}(\boldsymbol{x}) \delta \boldsymbol{x}$ for infinitesimal perturbations $\delta \boldsymbol{x}$. The system $\dot{\boldsymbol{x}} = f(\boldsymbol{x}, \boldsymbol{u})$ is contracting with rate $\lambda > 0$ if there exists a uniformly positive definite metric $\boldsymbol{M}(\boldsymbol{x})$ such that the Jacobian of the dynamics satisfies a matrix inequality:
\begin{equation}
\begin{aligned}
{\dot{\boldsymbol{M}}(\boldsymbol{x})}
&+ \frac{\partial f}{\partial \boldsymbol{x}}(\boldsymbol{x}, \boldsymbol{u})^{\top}
\boldsymbol{M}(\boldsymbol{x}) \\
&+ \boldsymbol{M}(\boldsymbol{x})
\frac{\partial f}{\partial \boldsymbol{x}}(\boldsymbol{x}, \boldsymbol{u})
\preceq -2 \lambda \boldsymbol{M}(\boldsymbol{x}),
\quad \forall \boldsymbol{x}, \boldsymbol{u}.
\end{aligned}
\end{equation}

{Here $\dot{\boldsymbol{M}}(\boldsymbol{x}) = \sum_i \frac{\partial \boldsymbol{M}}{\partial x_i}\, f_i(\boldsymbol{x},\boldsymbol{u})$ denotes the derivative of the metric along the system flow. This term must be included for a state-dependent metric $\boldsymbol{M}(\boldsymbol{x})$, and the commonly quoted form without it is recovered only in the special case of a constant metric $\boldsymbol{M}$.} This condition means the distance between two nearby solutions measured by $\boldsymbol{M}$ decays at least at rate $\lambda$. One can interpret it as a generalized Lyapunov condition applied to the difference dynamics. If we consider the error $|\tilde{\boldsymbol{x}}|= \boldsymbol{x}_1 - \boldsymbol{x}_2$ between two trajectories $\boldsymbol{x}_1$ and $\boldsymbol{x}_2$, contraction theory guarantees $|\tilde{\boldsymbol{x}}| \le k e^{-\lambda t} |\tilde{\boldsymbol{x}}(\boldsymbol{0})|$ for some $k>0$. A remarkable property is that if a system is contracting, it has a unique equilibrium or invariant trajectory that is globally exponentially stable \citep{lohmiller1998contraction}. In control contexts, we often speak of control contraction metrics (CCMs). A CCM extends the problem to controlled systems. It provides a condition under which a feedback controller can be designed so that the closed-loop system is contracting. If a CCM exists, one can synthesize a controller that achieves exponential tracking of any feasible reference trajectory \citep{manchester2017control}.

Contraction metrics are particularly useful in scenarios requiring robust trajectory tracking and disturbance rejection. A major challenge of this approach is finding a contraction metric for a given system, which can be challenging as finding a Lyapunov function. For complicated robots, $\boldsymbol{M}(\boldsymbol{x})$ might need to be very complex or even non-existent in closed form \citep{giesl2021computing, giesl2022review}. Another challenge is that contraction typically addresses stability properties but not explicit state constraints \citep{singh2017robust}.

\textbf{Reachability analysis and invariant sets} approaches compute the set of states from which the system can remain safe indefinitely or the set of states that could lead to an unsafe condition. Techniques like Hamilton-Jacobi (HJ) reachability solve a PDE to find the backward reachable set of an unsafe target \citep{mitchell2005time, bansal2017hamilton}. Consider a system whose dynamics are described by an ordinary differential equation
    $ \dot{\boldsymbol{x}} = \boldsymbol{f}(\boldsymbol{x}(t), \boldsymbol{u}(t)) $
where $\boldsymbol{x}$ is the system's states, $\boldsymbol{u}(t)$ is the control input.
To compute the reachable set, {a value function $V(\boldsymbol{x}, t)$ is defined whose zero sublevel set $\{\boldsymbol{x} : V(\boldsymbol{x},t)\le 0\}$ is the backward reachable set of the unsafe (avoid) target, the set of states from which the system cannot avoid reaching that unsafe target within the time horizon, so that the safe set is its complement.} This value function satisfies the HJ equation:
\begin{equation}
    \frac{\partial V}{\partial t} + H(\boldsymbol{x}, \nabla{V}) = 0
\end{equation}
where $H$ is the Hamiltonian defined as
\begin{equation}
    H(\boldsymbol{x}, \nabla{V}) = \max_{u \in \mathcal{U}}\{\boldsymbol{f(x, u)} \cdot \nabla{V}\}
\end{equation}
{The optimization over $\boldsymbol{u}\in\mathcal{U}$ in the Hamiltonian encodes the control for the chosen avoid formulation, following the convention of \citep{bansal2017hamilton, mitchell2005time}. Intuitively, it propagates the unsafe target backward in time so that the zero sublevel set $\{\boldsymbol{x} : V(\boldsymbol{x},t)\le 0\}$ collects the states from which the unsafe set cannot be avoided, and safety is maintained by keeping the system in its complement.}
By solving this equation, the boundary of the reachable set can be obtained, allowing for the analysis of the system's behavior under various control strategies \citep{mballo2025hamilton, choi2023forward, liu2024hamilton}.

The limitation is the curse of dimensionality. Solving HJ PDEs or exhaustive reachability is only feasible for simplified systems \citep{darbon2016algorithms, bansal2021deepreach, ganai2023iterative}. In safe learning contexts, reachability is often used as a runtime safety filter, where the reachable set boundary serves as a constraint. The filter minimally adjusts the nominal control input to satisfy the safety condition, thereby guaranteeing forward invariance of the safe set without strictly limiting the learning agent's performance \citep{liu2025reachable, 271, selim2022safe, kochdumper2023provably, suh2025dexterous}. 

\begin{figure*}[ht]
    \centering
    \includegraphics[width=0.95\linewidth]{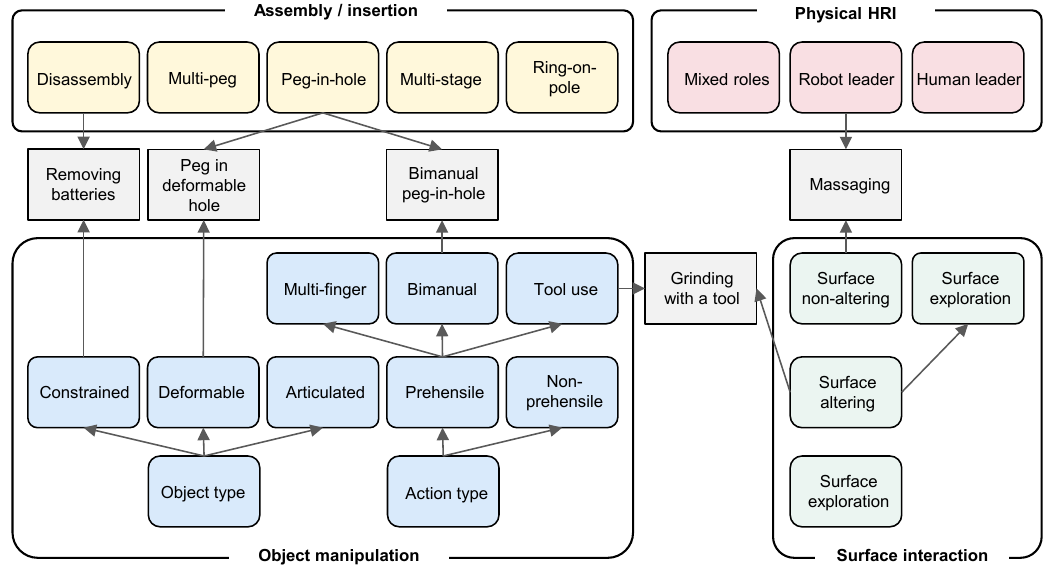}
    \caption{We review the contact-rich tasks under four main categories indicated with separate colors. Each category is further divided into sub-categories. The categories are not mutually exclusive. Examples of some downstream tasks that belong to multiple categories are shown as rectangular nodes.  }
    \label{fig:task-categories}
\end{figure*}


In this section, we introduce the \textbf{contact-rich task categories} from a safety-concerned online learning methods perspective. 
In accordance with our scope, we compile and discuss the contact-rich robotics tasks that are tackled with these methods in the literature. 
For this purpose, we identified the tasks in each of the papers we scanned for this survey. 
Our analysis shows that the \textit{insertion} or \textit{assembly} tasks are the most common type of physical interaction tasks. These are followed by the \textit{surface interaction} tasks such as polishing and grinding. 
We also identify the task types of \textit{object manipulation} and \textit{physical human-robot interaction (pHRI)}. 
We reviewed the tasks in the recent works under these four categories as shown in Fig.~\ref{fig:task-categories}. Details and downstream examples of these categories are given in the following sections. Please see Fig.~\ref{fig:task-images} for example task illustrations.

Our categorization is application-independent, however, it is possible to categorize the tasks also w.r.t. the application areas such as industry, household and healthcare \citep{Tsuji2025asu}.
Many contact-rich tasks are demonstrated for household applications that cannot be automated, as opposed to industry. These include general household tasks such as wiping a table or a vase \citep{151, 6}, inserting a plug \citep{112, 146}, and opening a door \citep{36, 149}; and kitchen tasks such as cutting vegetables \citep{21, 157}, stirring or scooping food \citep{92, 191}, and rolling pizza \citep{156}.
Industry applications usually focus on flexible manufacturing tasks such as inserting a gear \citep{177, 192}, nut threading \citep{noseworthy2025forge}, building furniture \citep{43}, and disassembly \citep{20, 62}. 
Healthcare applications mostly include the surgery operations \citep{339, 188} and physically assisting humans. 

Please note that there may be overlaps and similarities between the four categories as shown in Fig.~\ref{fig:task-categories}. 
For example, many peg insertion tasks could also be seen as object manipulation, or specific tool use tasks may aim for surface interaction. 
Depending on the problem formulation some downstream tasks may bring the challenges of multiple categories, which are detailed below.

\noindent\textbf{Assembly and insertion}\label{sec:task-assembly}
tasks have been studied in the control literature for a long time both for their importance in industry and the challenges they offer \citep{jiang2020state}. 
Robotic solutions to this problem can enable flexible assembly applications \citep{67}, and human-like activities such as inserting a key or a plug \citep{179, 85}.

The challenges include the precision requirement to achieve the tight fit, safety requirements to avoid damaging the equipment or getting stuck, and sensing difficulties as {the vision} modality is less informative due to occlusions and the force-based nature of the task. 
Assembly problems may involve multi-step insertion actions to achieve a more complex plan \citep{43, noseworthy2025forge}. 
The classical peg-in-hole task focuses on a single insertion of a securely grasped peg into a single upright hole \citep{levine2014learning}. 
The problem is gradually made more challenging through limited or noisy sensor data, leading to uncertainties over the peg and hole poses \citep{104}, varying peg shapes and sizes \citep{87}, or limited control options, such as rigid position-control~\citep{96}.

Other varieties were introduced for covering different application cases: 
Inserting multiple pegs (e.g., plugs) simultaneously \citep{30, 72}, 
assembly in obstructed environments that also require path planning \citep{99}, 
bimanually manipulating the hole and peg in the same time \citep{12, 31}, 
multi-stage insertion of vertically complex pegs (e.g., L-shape) \citep{27}, 
inserting into a deformable hole \citep{193}, 
reverse insertion (ring-on-pole) \citep{192}, 
and bolt threading \citep{noseworthy2025forge}.
The task varieties add new challenges such as synchronization between manipulators, added degrees of freedom, added constraints such as obstacles, multi-stage task solving and more complex dynamics.

Additionally, the \textit{disassembly} tasks should be mentioned here.
They invert the assembly task goal: There is already an inserted piece that has to be removed such as extracting a tightly fit part \citep{62}, sliding a bolt along a groove \citep{20}, or removing batteries \citep{113}.
The challenges of the disassembly tasks are similar to constrained object manipulation tasks such as opening a door or a drawer. However, disassembly presents tighter constraints that may easily lead to jamming.



\textbf{Surface interaction tasks} involve continuous and purposeful physical contact with a surface throughout task execution.\label{sec:task-surface} They play an important role in both industrial applications that are hazardous to humans, such as sanding~\citep{89}, and everyday assistive tasks in human environments, including table cleaning~\citep{35} and porridge stirring~\citep{191}.
These tasks usually require keeping a stable contact with the surface and applying consistent force while also following a desired trajectory. Maintaining a precise force and velocity profile enables the consistent quality of results. The problem is made more challenging by varying the surface curvature \citep{2, 142} or tracking complex motion trajectories \citep{107}. The robot may be equipped with a power tool such as a polisher that causes significant vibrations \citep{425online}.

The goal of the surface interaction tasks is to achieve a change on the surface. We observe these tasks under 3 categories: \begin{enumerate*}
    \item \textit{Surface altering} tasks that change the surface geometry by removing material, such as grinding and sanding \citep{89, 142};
    \item \textit{surface non-altering} tasks that do not cause a physical change in the surface, such as wiping and polishing \citep{66, 56};
    \item \textit{particle interaction} tasks that go beyond the surface and move particles, such as scooping a soft tissue \citep{188} or stirring a granular medium  \citep{92, 191}. 
\end{enumerate*}
These tasks may require intricate 6-dof maneuvers and high force to move through or process the particles \citep{12}.
In case of deformable objects \citep{zhu2022challenges}, \textit{surface altering} is easily achieved and revertible; in case of elastic objects such as a sponge \citep{44}, shape deformation is temporal and naturally reversed. Massaging is a special surface interaction task that is challenging due to involving a human, thus the safety concerns are more critical \citep{46}.
Interaction with deformable surfaces is an under-explored area, attracting more contributions.

We also identify the \textit{surface exploration} tasks as a type of surface non-altering tasks that are less explored in the literature.
In these works, the touch is fundamental to facilitate navigation and direct the agent towards the goal, such as in blind maze exploration \citep{zhang2024srl, 375}.
These tasks are similar to other surface interaction tasks as the interaction is usually non-prehensile and continuous, however, the goal of these tasks is not manipulating the surface, but navigating or following a trajectory \citep{90}. 

Surface interaction tasks are different than the object manipulation tasks as the target object is not expected to move.
On the contrary, it is desired to keep the object stable while achieving the surface interaction. 
The object of these tasks are usually physically constrained, however, wiping the surface of an unconstrained object, such as a vase, requires bimanual coordination to keep it stable during interaction \citep{6}.

\textbf{Object manipulation tasks} involve sustained and dynamic physical interactions with one or more objects throughout task execution.\label{sec:task-object} In most cases, the manipulated object itself is the target of the task. However, objects may also be grasped and used as tools to perform physical interaction tasks~\citep{196}. Such tools enable indirect interaction with the environment and can facilitate surface interaction, insertion, and other contact-rich manipulation tasks.

The object manipulation tasks are commonly categorized as prehensile and non-prehensile. The former requires grasping an object, while the latter interacts with the object without holding it, such as pushing \citep{94, 99}, sliding \citep{81} and pivoting \citep{145}. 
Non-prehensile tasks can be done with a simpler manipulator that maintains a single contact with the object, however, they also have a limited control over the object motion. Consequently, the safety of these tasks depend on the motion of the object in challenging environmental conditions: \cite{99} evaluate their method in cluttered environments where collisions are frequent and the object should be moved through narrow paths; the experiments of \cite{94} involve pushing the object through a narrow gap where the object orientation becomes important, or through a bridge with the risk of the object falling. 
On the other hand, prehensile tasks require at least two contacts with the object to maintain a grasp. 
Simple grippers can tightly grasp an object, however they do not allow complex object manipulations, thus they are limited to pick-and-place or tool-use tasks.
In contrast, bimanual manipulation is done through two high-DoF manipulators, and it can achieve complex interactions with the object such as lifting, pushing, flipping and rotating \citep{22, 53}.
Safe bimanual manipulation requires maintaining safe levels of contact forces without dropping the object, in addition to the workspace and motion limits \citep{22}.
Multi-fingered hands allow high manipulability of the object with increased control complexity. For a detailed review on dexterous hand manipulation we direct the reader to another survey \citep{huang2025human}.

The object manipulation tasks can be further categorized {with respect to} the nature of the object, such as constrained, articulated, and deformable.
Constrained object manipulation tasks, such as opening a door or a drawer, impose strict kinematic constraints on the object motion \citep{86}. These constraints may lead to high levels of force, if not handled with compliant and adaptive behavior \citep{36}. Objects like a door, drawer or lever are also termed articulated objects \citep{7}. 
\textit{Disassembly} tasks are challenging examples of constrained object manipulation. 
On the other hand, deformable object manipulation is an under-explored research problem that is challenging due to the sensing and modeling complexities \citep{zhu2022challenges}.

\textbf{Physical human--robot interaction (pHRI) tasks} require robots to intentionally establish and maintain physical contact with humans, in contrast to human--robot coexistence scenarios, where safety is primarily achieved by avoiding human contact~\citep{52, 57}.\label{sec:task-hri} These tasks present unique challenges because they must simultaneously guarantee human safety and accommodate the inherent uncertainty, variability, and unpredictability of human behavior during physical interaction.
Thus, the methods should estimate the human behavior and accommodate the uncertainty \citep{ajoudani2018progress}. 
In traditional pHRI, the robot assumes the passive role, and follows the human intention through compliance control. 
A special case of this is a robot interacting with both the human and the environment, as in \cite{17}, where the human guides the robot while it interacts with a surface.
In \cite{59}, the robot can switch between human-guided and autonomous in-contact manipulation modes to adapt its impedance characteristics.

The safety-critical contact-rich interactions between an active robot and human fit better to our scope. 
In prominent scenarios, the robot touches the human while the human takes a passive role, such as eating \citep{337}, dressing \citep{9} and bathing assistance \citep{5}, ultrasound scanning \citep{14}, and massaging \citep{19, 46}.
When the robot is physically coupled with the human, such as the rehabilitation robotics, the controller should estimate the human impedance parameters to ensure stability \citep{9531394}.

Crucially, the notion of safety expands with the inclusion of a human. For example, \cite{5} limit the vision system to a LIDAR during bathing assistance for ethical considerations. Human \textit{perceived safety} also becomes important for the trustworthy and socially acceptable interactions \citep{bartneck2009measurement}. 

\begin{tcolorbox}[
  colback=gray!3,
  colframe=gray!50,
  title={\Large\textcolor{black}{\textbf{Takeaway: Safety Across Contact-Rich Tasks}}},
  fonttitle=\bfseries,
  left=1mm, right=1mm, top=1mm, bottom=1mm
]
{
The safety notion depends on the type of physical interaction. \\
In \textbf{assembly and insertion}, safety is closely tied to avoiding jamming, excessive contact forces, and damage under tight geometric constraints. \\
In \textbf{surface interaction}, the challenge shifts toward maintaining stable, purposeful contact while regulating force, velocity, and trajectory quality over extended interactions. \\
\textbf{Object manipulation} introduces additional risks through object motion, grasp stability, environmental contacts, and deformable or constrained dynamics. \\
In \textbf{pHRI}, safety becomes even broader, incorporating not only physical injury prevention but also uncertainty in human behavior, ethical sensing choices, and perceived safety. 
While many of these task families have long been studied in robotics, recent work continues to introduce less constrained and more diverse scenarios, including deformable contacts, multi-stage interactions, cluttered environments, and human-centered applications. 
}
\end{tcolorbox}

\subsection{Sensing and Policy Modalities}}

In contact-rich manipulation, the choice of policy input has a direct impact on safety. The controller must detect the onset and evolution of contact, respect safety constraints such as joint/torque limits, admissible contact forces and friction cones, and remain robust to modelling errors and partial observability. 
To this end, modern safe learning pipelines leverage a spectrum of sensing modalities, ranging from purely proprioceptive state, through force/torque and high-bandwidth vision, to rich multimodal and language-conditioned inputs. 
Table~\ref{tab:inputs} summarizes the dominant input types and how they are typically exploited for safety.

\begin{sidewaystable}[t]
      \centering
      {
      \footnotesize
      \renewcommand{\arraystretch}{1.25}
      \caption{{Structured cross-modality comparison of sensing and policy inputs for safe contact-rich learning. For each modality we give its typical safety leverage and then contrast which safety quantity it can {directly bound} versus only {infer}, the resulting guarantee character, its sensing and modeling requirements, its cost and
  bandwidth, and its principal failure mode. The supporting references for each modality are given in the text of this section.}}
      \label{tab:inputs}
      \newcolumntype{Y}[1]{>{\hsize=#1\hsize\RaggedRight\arraybackslash}X}%
      \begin{tabularx}{\textwidth}{@{}>{\raggedright\arraybackslash}p{1.7cm}
    Y{1.0385} Y{1.0385} Y{0.8077} Y{1.2692} Y{0.8077} Y{1.0385}@{}}
      \toprule
      \textbf{Modality} & \textbf{Typical safety leverage} & \textbf{Safety quantity} & \textbf{Guarantee character} & \textbf{Key requirements / assumptions} & \textbf{Cost / bandwidth} & \textbf{Principal limitation}\\
      \midrule
      Proprioceptive / kinematic state & Contact reasoning from joint configurations and end-effector pose, conservative motion generation & Kinematic safety {directly}: joint/velocity limits, collision distances, geometric invariant sets & Configuration-space, model-based reasoning, no contact-force certificate & Accurate kinematic model and calibrated pose, object/obstacle poses observable, good planner & Low-dimensional, cheap, high-rate, no contact hardware & Cannot bound interaction force/pressure, blind to unmodeled/unexpected contact, only as safe as the geometric model\\

      Force/torque & Real-time enforcement of force and impedance limits, slip detection, compliant correction of model error & Interaction wrench {directly}: contact force/torque, impulse, impedance/passivity energy, slip/jamming onset & Direct measurement-based bound, supports hard force caps and passivity/stability certificates from the measured wrench & Wrist/finger F/T sensor with bias/gravity calibration, torque-controlled or compliant robot & Cheap sensor, low-dimensional, high control-rate, negligible compute & Local and non-anticipatory, missing distributed-contact detail\\
  
      Vision & Contact evaluation, localizing task features, and supplying global context that force sensors cannot provide & Scene geometry, object/obstacle pose, human proximity & spatial risk maps and predicted contact, not a measured-force bound & Calibrated cameras, adequate lighting, occlusion-free view, pose-estimation or learned perception & High-bandwidth, heavier perception compute and latency & Cannot directly bound contact force/pressure, degrades under occlusion, poor lighting, or pose-estimation error\\
      
      Natural-language instruction & Large multimodal models interpret a human's instruction and then check internal low-level safety & Semantic/task-level intent and constraints only, {no physical quantity} & Semantic-only: specifies and grounds constraints but cannot itself certify physical safety, a downstream filter is needed & grounding to symbolic plan, constraint set / safety filter for verification & High inference latency, off the real-time loop, coarse update rate & No measurement of force/geometry, can hallucinate or mis-ground, physical safety still rests on low-level controllers\\
      
      Tactile & High-resolution contact localization, pressure distribution monitoring, and incipient slip detection & Local contact state {directly}: pressure distribution, shear, contact location, incipient slip, bounds per-taxel/contact-patch loads & Direct, high-resolution local measurement, inherently local & Tactile skin / vision-based fingertip at the contact sites, calibration / representation learning & High-resolution data but spatially local, medium rate & Senses only where it touches, no global context, coverage and durability limited, cannot anticipate non-local events\\

      Multimodal fusion & Enables high precision under occlusion & Combined: global geometry, bounded local interaction, and intent in a shared representation & Composite: anticipatory {where}-to-contact plus measurement-based {how-hard} bounds from F/T and tactile & Multiple calibrated sensors with temporal/cross-modal alignment, fusion architecture, higher data/engineering burden & Highest aggregate bandwidth and compute, synchronization overhead & Complexity and calibration burden, modality dropout/miscalibration can mislead, no bound unless a measuring modality is present\\
  
      \bottomrule
      \end{tabularx}
      }
  \end{sidewaystable}

\textbf{Pose and proprioceptive state information} forms the basis of many safe learning approaches, which rely predominantly on proprioceptive and kinematic observations, including joint configurations, end-effector poses, and their derivatives, sometimes augmented with a small number of task-level geometric variables, while omitting explicit force/torque (F/T) measurements from the policy input.\label{sec:modality-proprioception} In these settings, safety primarily emerges from conservative motion generation, collision-aware planning, and geometric reasoning about contact, often supported by explicit constraints, recovery sets, or safety-oriented value functions~\citep{21, 22, 27, 34, 42, 45, 48, 91, 96, 99, 103}.

A first group of approaches couples a compact state representation (pose and velocity) with an explicit description of goals and safety constraints, and then uses model-predictive or safe-RL updates to keep trajectories inside a certified safe set. The policy or controller observes the current system state together with reference poses and bounds on states and inputs, sometimes augmented with uncertainty estimates, safety penalties or human-provided safety signals \citep{21, 22, 42, 45, 54, 57, 103}. Because distances to constraint boundaries are part of the observation, the algorithm can reason explicitly about safety margins without having to infer risk from raw contact forces, e.g. tightening constraints near obstacles, slowing down near joint limits, or triggering backup controllers when the state approaches a learned unsafe region.

A second family of methods organises behaviour around pose-based skills or motion primitives. Here, the observation typically consists of robot joint or end-effector states together with task-relevant geometric features, such as relative part poses or planner-generated waypoints, while the policy outputs bounded pose increments or selects from a library of safe trajectories \citep{27, 34, 96, 99}. Motion-planner-augmented RL and planner or demonstration-driven frameworks exemplify this pattern. The learned policy reasons over kinematic state and high-level task features, whereas collision-free paths and smooth accelerations are enforced by a downstream planner or stabilising controller. Safety is thus embedded in the way pose sub-goals are parameterised and filtered, ensuring that abrupt or large displacements are routed through a geometry-aware planning layer.

Pose-dominant observations are also used when exteroceptive information is represented directly in configuration space rather than as raw pixels. Benchmark interaction tasks encode the positions of pucks, tools or human partners as Cartesian poses, combined with the robot’s own pose and velocity, so that learning algorithms can enforce distance-based constraints, safety fields around humans, or recovery zones in a low-dimensional geometric space \citep{91, 48, zhang2025bresa}. Even in visuomotor settings where images are available, the safety shaping is often defined over the proprioceptive and pose coordinates through joint limits, collision distances or certified invariant sets, while the visual input mainly serves to infer these geometric quantities or to disambiguate task context \citep{13, 47, 95, 99, 107}. Compared with force-centric designs, such pose-based formulations trade direct measurement of contact forces for simpler, more structured safety reasoning in configuration space, and are particularly attractive when high-quality kinematic models and planners are available.

\textbf{Force and torque sensing} provides the primary safety interface for many contact-rich manipulation tasks, particularly insertion, surface following, and physical human--robot interaction.\label{sec:modality-ft} Policies typically combine force/torque (F/T) measurements from the wrist or fingers with joint and end-effector states to regulate contact loads, detect slip or jamming, and adapt impedance so that tracking performance is maintained without violating force or pressure limits~\citep{zhang2024srl, 1, 2, 4, 23}. Compared with purely kinematic formulations, these approaches expose safety-critical quantities directly in the observation space, enabling controllers to respond immediately to unsafe contact events in assembly, surface interaction, and terrain-contact scenarios~\citep{19, 32, 39, 61}.

A large family of methods couples F/T sensing with variable impedance or admittance control. The policy input in these works typically concatenates joint or end-effector pose and velocity with measured interaction forces/torques, and sometimes task descriptors such as surface or tissue properties, environment stiffness, or task phase \citep{2, 19, 32, 46}. The action space is parameterized in terms of stiffness, damping, or target forces, and safety is enforced by bounding these impedance parameters, shaping them over time, or embedding stability and performance objectives into the learning problem \citep{23, 25, 41, 78}. This pattern appears across industrial assembly and sanding, massage and rehabilitation, and sliding or surface following tasks, where the policy learns to reshape compliance in response to the measured wrench so that interaction forces remain within prescribed envelopes \citep{39, 61, 69, 71}. Some approaches further introduce latent policy classes or structured action spaces over impedance parameters, which makes it easier to impose smoothness and magnitude limits on the learned stiffness and damping profiles \citep{1, 40, 59, 74}.

F/T signals are also combined with explicit safety margins, constraints or stability certificates derived from control theory. In this line of work, the observation space does not stop at raw wrench measurements and state. Controllers additionally receive features derived from Lyapunov functions, control barrier functions, passivity errors or other safety indicators that are computed using the measured forces and state trajectories \citep{14, 16, 29, 59}. Learning algorithms then operate directly on these structured safety variables, for example by penalizing trajectories that approach constraint boundaries, adjusting admittance or impedance gains when passivity residuals grow~\cite{zhang2025passivity, zanella2024learning}, or biasing exploration away from regions with high estimated risk \citep{39, 44, 57, 80}. This idea extends naturally to uncertainty-aware formulations, where system state, force/torque and uncertainty estimates are combined, and the policy is trained to remain robustly within contact-safe sets despite modelling errors and partial observability \citep{60, 92}.

In physical human–robot interaction and teleoperation, F/T channels additionally encode human intent and comfort. Many frameworks treat human-applied forces at the end-effector, interaction torques, or teleoperated wrenches as part of the policy input together with robot joint or end-effector state and task phase \citep{11, 17, 9531394, 58}. These signals are interpreted either as demonstrations of desired compliant behaviour or as real-time preferences over interaction forces, and are used to adapt impedance online, to learn tele-impedance mappings, or to construct task-oriented safety fields around the human partner \citep{15, 52, 69, 76}. Safety here is twofold: the robot must both respect human-specified force envelopes and ensure that the resulting closed-loop behaviour remains stable under changing human dynamics.

Another recurrent pattern is to use F/T sensing for contact-state estimation, search, and long-horizon assembly. In such settings, policies augment pose and geometry with force histories, contact-state labels, or belief maps over likely contact locations \citep{3, 7, 26, 55}. Contact cues extracted from wrench measurements, such as contact onset, sticking versus sliding, or successful insertion, are then used to gate skills, switch admittance gains, or trigger recovery behaviours when jamming or misalignment is detected \citep{29, 32, 37, 72}. Active-search methods in particular combine end-effector pose and force/torque with Bayesian belief updates and safety features, enabling the robot to probe cautiously, back off when measured forces exceed certified bounds, and refine subsequent search motions while remaining within safe contact regions \citep{14, 56, 60, 77}. Compared with vision-only pipelines, these force-centric designs provide a direct window into the quality of contact, enabling safer behaviour under occlusion, poor lighting, or inaccurate geometric models.

Even when richer sensory modalities such as vision or tactile arrays are available, force/torque remains a core channel through which safety is enforced. Multimodal policies often use vision primarily to infer geometry and high-level task context, while relying on force/torque and proprioception to regulate local interaction \citep{9, 12, 31, 94}. End-to-end sensorimotor models that consume images, joint state and force/torque typically define safety shaping, through penalties, auxiliary critics or recovery policies, over wrench and configuration features rather than pixels, underscoring that F/T sensing is the main substrate on which safe contact behaviour is learned and executed \citep{37, 95, 100}.

\textbf{Vision sensing} provides non-contact, global information about objects, surfaces, and humans that is difficult or impossible to infer from proprioception and force sensing alone.\label{sec:modality-vision} Depending on the camera configuration, visual observations range from bird's-eye or third-person views covering the entire workspace, to fixed eye-to-hand cameras monitoring a local workspace, and wrist-mounted or eye-in-hand cameras that continuously observe the interaction region from the robot's perspective~\citep{3, 5, 13, 15, 20, 36, 57, 65}. Across these configurations, vision contributes to safety in two complementary ways: by anticipating future contact and potential collisions through scene understanding and geometric reasoning, and by monitoring contact evolution and interaction dynamics in the immediate vicinity of the end-effector.

Global and external viewpoints are typically used to capture task geometry, object and fastener poses, and human motion. Assembly and disassembly systems exploit RGB or RGB-D images, often converted into scene graphs or object pose estimates, alongside joint and end-effector state to ensure that motion plans respect part tolerances and obstacle constraints before contact is made \citep{20, 36, 67, 95, 98}. Surface-interaction and bathing applications reconstruct patient or workpiece shape from point clouds or depth maps and use this geometry, together with robot pose, to generate reference paths that avoid sharp curvature and maintain safe approach angles during contact \citep{1, 2, 5, 61}. In human–robot collaboration, visual tracking of human pose or trajectories is fused with robot state and sometimes interaction forces to construct task-oriented safety fields around the person. These fields enter the policy input as spatial risk maps that reshape robot motion to avoid unsafe proximity or contact \citep{15, 57}. Pixel-based safe RL and visuomotor world models further push this idea by using raw images as the primary observation, with safety-aware critics or constraint penalties guiding behaviour toward image regions that correspond to safe configurations \citep{13, 22, 65, 91}.

Local eye-in-hand or wrist-mounted cameras provide a complementary, close-range view that is particularly valuable in contact-rich tasks where the relative pose between tool and environment must be controlled with high precision. Insertion and fastening pipelines use visual estimates of peg–hole or object–fixture pose from such cameras as part of the policy input, so that approach and alignment motions are corrected before large contact forces occur \citep{3, 28, 85, 87}. Diffusion-based compliance policies and force-constrained dressing controllers condition on sequences of RGB or RGB-D images from wrist or nearby cameras, together with end-effector poses, to predict how candidate motions will influence contact geometry and thereby avoid unsafe configurations \citep{6, 9, 12}. In variable-impedance manipulation with image-conditioned action spaces, multiple RGB views of the workspace are combined with end-effector pose and velocity, so that the controller can adapt stiffness and damping not only to measured forces but also to visually perceived contact conditions, occlusions and environmental changes \citep{89, 100, 107}.

A prominent trend is to fuse visual information with haptic and proprioceptive sensing in a joint representation that directly feeds safety-aware policies. Visual–haptic variable impedance controllers combine latent visual features extracted from RGB images with force/torque or tactile readings and robot state, enabling the policy to reason simultaneously about global geometry and local contact quality when selecting compliant actions \citep{31, 86}. Dressing and bathing systems integrate segmented point clouds or image-based features with end-effector force histories and candidate motions, using learned dynamics or prediction models to filter out actions whose predicted contact forces exceed task-dependent thresholds \citep{5, 9, 12}. In furniture assembly and other complex contact-rich tasks, multimodal context sequences concatenate proprioceptive vectors, RGB observations and high-resolution tactile readings, allowing residual policies or skill-transfer mechanisms to refine motion precisely at the moment of contact while keeping interaction forces within safe ranges \citep{38, 43, 82}. Recurrent or belief-state architectures extend this idea by maintaining a latent estimate of the scene and contact state under partial observability, with both vision and force/torque streams contributing to belief updates and safety-aware action selection \citep{94, 100}.

Finally, several frameworks place visual perception within larger safety-aware learning stacks that combine high-level reasoning with low-level control. Safety-aware unsupervised skill discovery and recovery-based RL methods treat vision as part of the state used to infer which regions of the environment are safe to explore, and to learn recovery zones or backup skills that can be invoked when the system approaches unsafe visual configurations \citep{22, 91, 94}. Vision–language–action models extend this further by taking images alongside natural-language instructions and past interaction histories, and by incorporating safety feedback into the training objective so that the resulting policies can interpret human intent while avoiding visually grounded unsafe behaviours \citep{33}. Overall, these works highlight that visual inputs, whether global or hand-centric and whether raw pixels or structured geometry, play a central role in making safety-relevant aspects of the environment explicit to the policy, so that contact-rich behaviour can be regulated proactively rather than reactively.

\textbf{Language instructions} provide an increasingly important high-level interface through which humans specify tasks, constraints, and preferences, while lower-level controllers remain responsible for enforcing physical safety during execution.\label{sec:modality-language} Rather than hard-coding task objectives into cost functions or reward formulations, these approaches allow users to describe what the robot should do, and often how cautiously it should behave, using natural language~\citep{kyaw_text_2025}. Such instructions are subsequently grounded into symbolic plans, motion goals, or policy parameters~\citep{8, 33, 51}. From a safety perspective, language complements geometric and haptic observations by explicitly communicating human intent and safety requirements at the policy interface.

One line of work couples language with visual and proprioceptive observations in large vision–language–action architectures. In such systems, camera images and robot state are processed jointly with a natural-language instruction to produce either low-level actions or intermediate code that can be executed by downstream controllers \citep{8, 33}. Safety is then imposed at two levels: the model is aligned to avoid generating obviously harmful plans, and the generated code or action sequences are checked against constraint sets or safety filters before deployment \citep{8, 33}. Compared to purely numeric task encodings, this design allows the same sensorimotor stack to be reused across tasks, while safety-relevant differences are captured in the language input rather than in separate controllers.

Another set of approaches emphasizes preference-driven interaction. Here, users do not only issue initial instructions, but also provide follow-up signals indicating whether a behaviour is safe, comfortable, or acceptable, and these signals are incorporated into the learning objective or alignment procedure \citep{33, 54}. Text-conditioned models, for instance, may take as input a description of the desired interaction style together with histories of states, actions and safety feedback, and are trained to predict actions that satisfy both task and safety preferences \citep{33, 51}. Human-centered safe RL frameworks complement this by treating human feedback as an additional observation channel that shapes the policy toward behaviours that humans deem safe, even when those behaviours are not easily captured by simple numerical constraints \citep{54}.

Across these works, language does not replace low-level safety mechanisms such as force limits, control barrier functions or conservative exploration. Rather, it provides a flexible interface for specifying and updating what safe means in a given context~\cite{33}. High-level instructions determine which regions of the state and action spaces are desirable or forbidden, while the underlying visuomotor and force-aware controllers ensure that, once a plan is chosen, contact forces, distances and stability constraints remain within acceptable bounds \citep{8, 33, 51, 54}. As language models and alignment techniques improve, this combination of linguistic intent with structured safety inputs offers a promising route to robots that can be both responsive to human guidance and reliably safe in contact-rich environments.

\textbf{Tactile sensing} provides direct measurements of contact states, including contact location, normal and shear forces, and incipient slip, making it a key sensing modality for the safe execution of contact-rich manipulation.\label{sec:modality-tactile} Among the various tactile sensing technologies, vision-based tactile sensors, such as the GelSight family~\citep{yuan2017gelsight, wang2021gelsight}, have become particularly popular in robot learning due to their ability to capture rich contact geometry and force-related information. In parallel, low-cost and scalable tactile sensing technologies, including flexible tactile sensor arrays~\citep{huang20243d}, have enabled dense tactile perception for contact-rich manipulation. By providing localized measurements of contact conditions that are unavailable from vision or proprioception alone, tactile sensing enables robots to detect slip, estimate contact quality, and regulate manipulation forces for safer physical interaction. There has also been progress in vision-based tactile sensors with task-adaptive geometries, such as hemispherical and finger-shaped designs \citep{11024242, sun2022soft, 9811966}. High-resolution vision-based tactile sensing provides high-resolution, local information about contact that complements global vision and low-dimensional F/T measurements. Rather than only measuring a single wrench at the wrist, tactile skins and fingertip arrays reveal pressure distributions, shear and incipient slip at the exact points of interaction, making it possible to detect unsafe contact patterns, such as concentrated loads on corners or sliding on fragile surfaces, before they escalate into damage or failure \citep{31, 43, 71}. Beyond conventional parallel-jaw grippers and tactile-equipped dexterous hands \cite{zhao2025ftac, bhirangi2023allthefeels, suresh2024neuralfeels}, other end-effectors—such as multi-DoF (e.g.,5-DoF) grippers—can enable manipulation primitives like cable/rope disentangling and thin-card flipping \cite{zhou2024hand}. These designs broaden the repertoire of feasible manipulation tasks and could be promising to combine with safety modules or learned control policies in future work. For safe learning, these rich contact cues are typically fed directly into the policy alongside proprioceptive state, so that both exploration and execution can be guided away from behaviours that produce sharp spikes or unstable contact transitions \citep{71, 106}.

One group of methods uses tactile arrays as the main channel for encoding task success and safe contact configurations. \cite{zhu2025touch} uses a large-scale visuo-tactile dataset consisting of over 2.6 million vision–touch pairs, collected across multiple environments and covering 43 manipulation tasks. The dataset is used to pretrain a visuo-tactile representation (encoder) as an upstream step, which can later serve as a foundation for downstream work such as safe contact-rich interaction or manipulation. In furniture assembly, for example, tactile ensembles distributed across fingers or grippers are used to characterise how surfaces, edges and fasteners behaviors during successful insertions or alignments. Policies are then trained to reproduce these tactile signatures while avoiding contact patterns associated with jamming or over-tightening \cite{43}. Gentle manipulation frameworks similarly treat touch as the primary feedback for regulating interaction intensity. Policies observe proprioception and tactile events, and are rewarded for acquiring task-relevant information while keeping per-taxel pressures and impact magnitudes low, which naturally steers exploration toward soft, sliding contacts rather than hard impacts \cite{106}. In both cases, safety is not an external constraint but an emergent property of how tactile patterns are encoded in the observation and reward.

Tactile sensing plays a key role in variable-impedance control for contact-rich motion along surfaces. In aerial or lightweight manipulators that slide along uneven terrain or structural elements, the policy often receives only proprioceptive state and tactile readings, without full base F/T sensing \cite{71}. The tactile stream reveals contact loss, stick–slip, and localized high-pressure/unsafe contact, while proprioception captures global motion; impedance can then be updated online—stiffening when contact is well supported and softening/backing off when tactile cues indicate high pressure or impending slip \cite{71}. Beyond aerial sliding, tactile/visuo-tactile interaction controllers explicitly use curvature/alignment monitoring to adapt stiffness and force regulation when exploring unknown 3D curved surfaces, including contact-loss recovery and passivity-aware safety mechanisms \cite{karacan2023tactile, karacan2024visuo}. For continuous surface tasks (e.g., cleaning/inspection), closed-loop tactile coverage methods encode the target on point clouds and use task-space impedance with force regulation to maintain contact while covering curved surfaces \cite{bilaloglu2025tactile}.

A further set of approaches embeds tactile input within multimodal visual haptic proprioceptive representations~\cite{364}. Hierarchical variable-impedance controllers, for instance, build high-level policies on visual features that capture object pose and scene layout, while lower-level modules refine actions using haptic channels that may include both force/torque and tactile measurements \cite{31}. In such designs, vision plans where and when contact should occur, while tactile feedback confirms whether contact is distributed and stable in the way the planner expects. More recent multimodal RL systems explicitly combine visual observations, tactile signals and joint state into a shared latent space, and train adaptive policies that allocate different roles to each modality. Vision is used to anticipate geometric constraints, tactile cues to detect unsafe local pressure or slip, and proprioception to enforce kinodynamic limits \cite{86}. By tying safety-related decisions to tactile cues rather than to coarse force thresholds alone, these methods achieve more nuanced and robust protection for delicate objects, complex assemblies and human collaborators.



\begin{tcolorbox}[
  colback=gray!3,
  colframe=gray!50,
  title={\Large\textcolor{black}{\textbf{Takeaway: Safety Roles of Different Modalities}}},
  fonttitle=\bfseries,
  left=1mm, right=1mm, top=1mm, bottom=1mm
]
{
No single modality dominates; the right choice follows the binding safety quantity.

\textbf{Force/torque sensing} is the primary substrate for bounding interaction forces and is hard to replace for force-limited tasks.

\textbf{Vision} contributes global, anticipatory context (approach, collision avoidance, scene geometry) but cannot by itself bound contact forces.

\textbf{Tactile sensing} supplies local pre-failure cues (incipient slip, pressure concentration) valuable for fragile contacts.

\textbf{Multimodal fusion} helps most under occlusion or inaccurate models.

The recurring effective pattern is therefore to use vision and language for \emph{where} and \emph{whether} to make contact, and force/tactile sensing with a compliant action abstraction for \emph{how hard}.
}
\end{tcolorbox}

\subsection{Data Acquisition}
\label{sec:data_acquisition}
Data acquisition plays a foundational role in safe learning for contact-rich manipulation. The quality, diversity, and safety-relevance of the data directly influence the performance and generalizability of learned policies, especially in scenarios involving unpredictable contacts or constrained physical interaction. In this section, we discuss three key paradigms in data acquisition: (i) simulation-based data generation, (ii) real-world data collection, and (iii) hybrid and self-supervised data acquisition.

\begin{figure*}[t]
    \centering
    \includegraphics[width=0.96\textwidth]{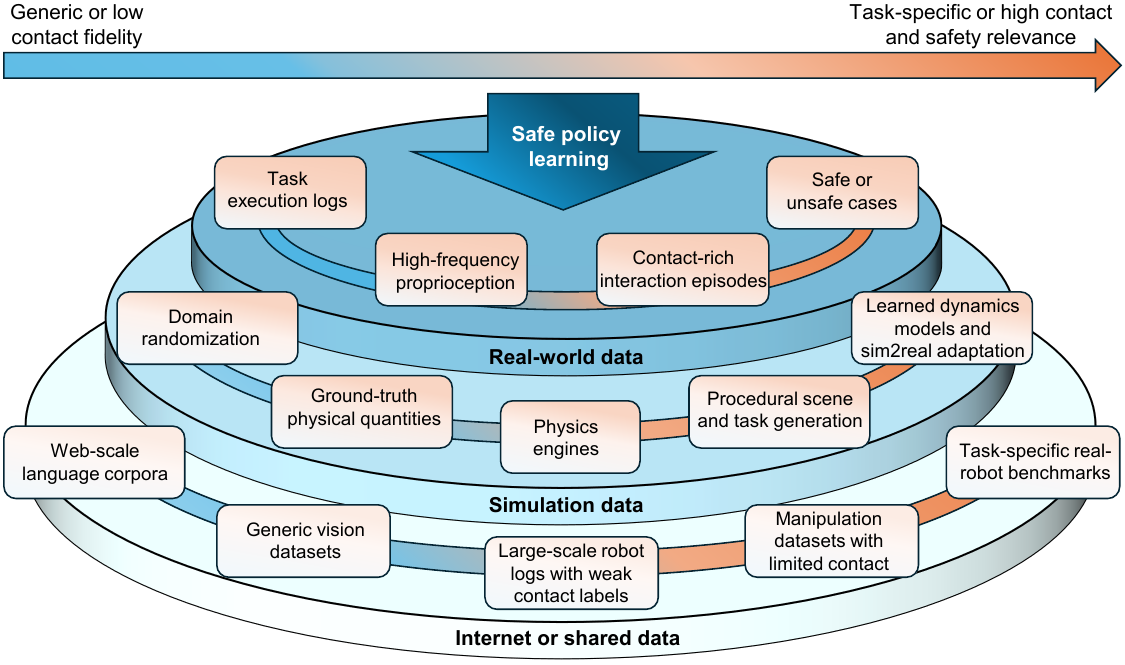}
    \caption{Data pyramid for safe policy learning in contact-rich manipulation. 
    The pyramid is organized into three layers according to physical fidelity and safety relevance. 
    (1) Real-world data: includes task execution logs, high-frequency proprioception (joint states, velocities, tactile and force--torque sensing ), contact-rich interaction episodes, and safe and unsafe or failure cases, which provide the most direct evidence about real physical contacts and safety-critical events. 
    (2) Simulation: includes domain randomization (dynamics, appearance, noise), ground-truth physical quantities (forces, poses, contacts), physics engines (MuJoCo, PyBullet, Isaac Lab), procedural scene and task generation, and residual or learned dynamics models with sim-to-real adaptation, enabling large-scale exploration of contact scenarios under controllable yet imperfect dynamics. 
    (3) Internet and shared datasets: includes web-scale language corpora (LLM data), generic vision datasets (web images and videos), large-scale robot logs with weak contact labels, manipulation datasets with limited contact or safety labels, and task-specific real-robot benchmarks with small scale, which are abundant but only indirectly capture contact-rich and safety-critical phenomena.}
    \label{fig:data_pyramid}
\end{figure*}

Data matters for policy learning, especially for a safe policy where negative samples play a significant role in training the agent to be aware of what unsafe scenarios are. However, acquiring such data is time-consuming and labor-intensive. Crucially, unlike the vast text and image corpora scraped from the Internet to train LLMs and VLAs, physical interaction data, specifically high-fidelity contact forces, torques, and safety-critical failure modes, cannot be obtained from the web. This scarcity of contact data in foundation models creates a significant gap in their ability to reason about physical dynamics and safety. Consequently, physical data collection remains inevitable and irreplaceable for contact-rich tasks, despite efforts to leverage simulation and internet-scale proxies. As shown in the data pyramid in Fig.~\ref{fig:data_pyramid}, real-world contact data sits at the apex and is both scarce and essential.
 And in Fig.\ref{fig:data_pyramid} we retain only the most representative concepts in each layer to highlight the progression rather than exhaustively listing all sources. Within each layer, the listed keywords are ordered by increasing contact fidelity and safety relevance: for real-world data, from generic task execution logs to explicit safety and failure cases; for simulation, from generic randomized environments to simulation elements that most directly support sim-to-real transfer; and for internet and shared datasets, from general web-scale corpora to task-specific real-robot benchmarks.

\textbf{Simulation-based data generation} provides a safe, scalable, and cost-effective means of collecting large volumes of training data for contact-rich manipulation without risking damage to robots, tools, or workpieces.\label{sec:data-simulation} Physics engines such as MuJoCo~\cite{todorov2012mujoco}, MuJoCo Playground~\cite{zakka2025mujoco}, PyBullet~\cite{coumans2016pybullet}, Isaac Gym~\cite{makoviychuk2021isaac}, and Isaac Lab~\cite{mittal2025isaaclab} enable learning agents to interact with objects under diverse contact conditions while providing ground-truth access to quantities such as contact forces, velocities, object poses, and other latent states that are difficult or expensive to measure on physical platforms. These simulators further support massively parallel rollouts, substantially accelerating the training of data-intensive methods, including deep reinforcement learning and imitation learning, for contact-rich robotic manipulation.

To improve generalization and robustness, domain randomization and procedural scene generation are widely used to expose policies to variations in geometry, mass, friction, sensor noise, and task parameters, which is crucial for sim-to-real transfer in assembly, insertion, and other tight-tolerance tasks~\cite{37, 139, 145}. Recent work further combines high-fidelity rendering and physics to generate multimodal datasets (vision, force/torque, tactile) that can be used to train VLM/VLA-style models and other foundation models for contact-rich robotics.

However, simulations still suffer from a pronounced reality gap, particularly for phenomena such as material deformation, frictional sticking and sliding, small-clearance contacts, and compliant or flexible components, which can lead to significant discrepancies in policy behavior when transferred to real hardware. To mitigate this, recent approaches incorporate learned residual models, online adaptation, or human-in-the-loop corrections on top of simulation-trained policies, for example by learning residual compliance parameters or gated residual policies that refine the base controller in the real world~\cite{111, 176}. Together, these trends point toward hybrid pipelines in which accurate simulation, structured domain randomization, and residual or adaptive components jointly address the sim-to-real challenges of safe contact-rich manipulation.

\textbf{Real-world data collection} remains indispensable for safe learning because it captures the complexities and uncertainties of physical interactions that cannot yet be faithfully reproduced in simulation.\label{sec:data-real} By deploying robots in controlled laboratory environments or real-world settings, researchers can collect data on contact forces, object dynamics, sensor measurements, and task performance under realistic operating conditions. Such data expose learning algorithms to unmodeled dynamics, sensing noise, hardware imperfections, and unexpected contact events, thereby improving the robustness and reliability of policies intended for deployment in safety-critical contact-rich tasks.

However, collecting high-quality data in the real world is often time-consuming, expensive, and potentially unsafe, especially in contact-rich tasks where excessive forces or unexpected contacts can damage hardware or pose safety risks to humans. To address these challenges, researchers often employ techniques such as teleoperation, where human operators guide the robot through tasks while collecting data on safe and unsafe interactions~\cite{mower2023ros, 354}. 
Additionally, safety constraints can be integrated into the data collection process, ensuring that the robot operates within predefined limits to avoid damaging itself or its environment~\cite{5, 18}.
Furthermore, real-world data can be augmented with simulation data through techniques like domain adaptation or transfer learning, allowing policies trained in simulation to be fine-tuned on real-world data~\cite{zhang2024srl, berrocal2024evaluating}.
\begin{table*}[h]
    \centering
    \small\setlength{\tabcolsep}{4pt}
    \caption{Representative simulation platforms used in learning-based contact-rich robotics, compared in terms of deformable-object support, GPU acceleration, and primary task domains. The table covers general-purpose physics engines, GPU-accelerated robot-learning platforms, and task-specific environments for contact-rich tasks.}
    \begin{tabular}{l|c|c|c|r}
    \hline
    \hline
      Simulators & Deformable & GPU & Tasks & References\\
      \hline
         MuJoCo & \checkmark & \xmark & Manipulation, Locomotion & {\cite{todorov2012mujoco}}\\ 
         MuJoCo XLA/MJX & \checkmark & \checkmark & Manipulation, Locomotion & {\cite{zakka2025mujoco}}\\ 
         Genesis & \checkmark & \checkmark & Soft-body, Articulated & \cite{Genesis}\\
         PyBullet & \checkmark & \xmark & Manipulation, Grasping & {\cite{coumans2016pybullet}}\\
         Isaac Gym & \checkmark & \checkmark & RL Training, Manipulation & \cite{makoviychuk2021isaac}\\
         Isaac Sim & \checkmark & \checkmark & Photorealistic, Digital Twin & \cite{noauthor_isaac_nodate, berrocal2024evaluating}\\
         Isaac Lab & \checkmark & \checkmark & Robot learning & {\cite{mittal2025isaaclab}}\\
         Orbit     &\checkmark & \checkmark & Interactive  Robot Learning &\cite{mittal2023orbit}\\
          
         Factory  &  \xmark & \checkmark & RL, Assembly & \cite{narang2022factory}\\
         RoboAssembly & \xmark&\xmark &furniture assembly & \cite{yu2021roboassembly}\\
         Drake & \xmark & \xmark & Model-based Control & \cite{tedrake2014drake}\\
    \hline
    \end{tabular}
    \label{tab:simulators}
\end{table*}

\begin{table*}[h]
    \centering
    \footnotesize\setlength{\tabcolsep}{3pt}
    \caption{Representative benchmarks commonly used to evaluate learning-based, contact-rich robotic systems. The comparison emphasizes scalability, deformable-object support, GPU acceleration, and core task domains, spanning general manipulation, reinforcement learning, and safety-oriented assessment.}
    \begin{tabular}{l|c|c|c|c|r}
    \hline
    \hline
      Benchmarks & Scalable& Deformable & GPU & Tasks & References\\
      \hline 
         RoboVerse & \checkmark& \checkmark & \checkmark & Manipulation, RL Benchmarks & \cite{geng2025roboverse}\\
         RoboCasa & \checkmark& \checkmark & \checkmark & Household Tasks, Manipulation & \cite{nasiriany2024robocasa}\\
         RoboSuite & \checkmark& \checkmark & \checkmark & Manipulation, RL Benchmarks & \cite{zhu2020robosuite}\\
         ManiSkill & \checkmark& \checkmark & \checkmark & Manipulation, RL Benchmarks & \cite{mu2maniskill, gumaniskill2}\\
         Robot Air Hockey & \checkmark & \xmark  & \checkmark & Air Hockey game & \cite{liu2024retrospective}\\
         Meta-World & \checkmark& \xmark & \checkmark & Manipulation, RL Benchmarks & \cite{yu2020meta}\\
         OpenAI Gym & \checkmark& \xmark & \checkmark & RL Benchmarks, Manipulation & \cite{brockman2016openai}\\ 
         RLBench & \checkmark& \xmark & \checkmark & Manipulation, RL Benchmarks & \cite{james2020rlbench}\\
         Stable Baselines & \checkmark& \xmark & \checkmark & RL Benchmarks, Manipulation & \cite{raffin2021stable}\\
         Gymnasium & \checkmark& \xmark & \checkmark & RL Benchmarks, Manipulation & \cite{towers2024gymnasium}\\
         Safety Gymnasium & \checkmark& \xmark & \checkmark & Safety in RL, Manipulation & \cite{152}\\
         Robust gymnasium & \checkmark& \xmark & \checkmark & Safety in RL, Robotics & \cite{gu2025robust}\\
        Safe-Control-Gym & \checkmark& \xmark & \checkmark & Safety in RL, Control & \cite{yuan2022safe}\\
            \hline
    \end{tabular}
    \label{tab:benchmarks}
\end{table*}

\textbf{Hybrid and self-supervised data acquisition} combines simulation, real-world experience, and autonomous data generation to improve the scalability and robustness of safe learning.\label{sec:data-hybrid} Hybrid strategies leverage the complementary strengths of simulation and physical interaction by pre-training policies in simulation and subsequently adapting them using real-world data, or by augmenting limited physical datasets with simulated experience~\citep{berrocal2024evaluating}. Such approaches enable efficient exploration of diverse contact scenarios while retaining the ability to adapt to the complexities of real-world interactions. Complementarily, self-supervised learning exploits the robot's own experience to generate additional supervisory signals without requiring extensive manual annotation~\citep{zhang2024srl}. For example, robots may learn to predict future states, contact forces, or interaction outcomes directly from multimodal sensory observations, thereby acquiring representations that improve policy robustness and exploration. Together, hybrid and self-supervised data acquisition strategies help bridge the sim-to-real gap, expand the diversity of contact-rich training data, and improve the generalization and safety of learned manipulation policies.

\begin{tcolorbox}[
  colback=gray!3,
  colframe=gray!50,
  title={\Large\textcolor{black}{\textbf{Takeaway: Trade-offs in Data Acquisition}}},
  fonttitle=\bfseries,
  left=1mm, right=1mm, top=1mm, bottom=1mm
]
{
Simulation provides a scalable and safe means of data collection, but the sim-to-real gap is particularly pronounced for safety-critical contact tasks, including friction, small-clearance interactions, and deformation. Real-world data capture these effects more faithfully but are costly and potentially risky to collect, particularly for safety-critical failures and near-miss events.\\
A practical strategy is therefore to combine large-scale simulation and domain randomization for broad coverage with monitored real-world fine-tuning or adaptation, leveraging the complementary strengths of both data sources.
 }
\end{tcolorbox}

\subsection{Simulation Environment and Benchmarks} \label{subsec:simAndBench}
Simulation plays a critical role in the development and evaluation of safe learning algorithms for contact-rich robotic tasks.
These tasks often involve complex dynamics, high-dimensional state spaces, and the potential for hardware damage, making direct deployment on real robots risky and costly during early stages of learning. 
Simulated environments provide a safe, controllable, and repeatable platform to test algorithms, enabling rapid iteration and the generation of large-scale data without physical wear~\cite{long2025survey}.

Moreover, standardized benchmarks enable consistent comparison across different approaches, thereby fostering reproducibility and progress in the field. By utilizing well-defined simulation platforms and task suites, researchers can more effectively assess safety, generalization, and performance under varied conditions before transitioning to real-world applications.

\textbf{Simulation environments} can be grouped into a richer taxonomy reflecting the underlying physics engine, parallelism, sensing realism, and intended evaluation protocols, as detailed in Table~\ref{tab:simulators}. We highlight the perspective of contact-rich tasks in this survey, while readers can find more comprehensive comparisons in~\cite{berrocal2024evaluating}.

\textbf{General robotics simulators (ROS-ready)} emphasize sensor realism (e.g., RGB-D and LiDAR), ROS integration, and extensible plugins for actuation and perception. Representative platforms include Gazebo~\cite{38, 70, 71, 84, 96, 175}, Webots~\cite{95}, CoppeliaSim~\cite{169}, and Drake~\cite{196}, making them well-suited for closed-loop, system-level validation, including the evaluation of safety monitors.

\textbf{Rigid-body dynamics engines} are primarily designed for accurate simulation of contact dynamics and articulated systems. \emph{\textbf{CPU-based engines}} such as DART (via robot\_dart)~\cite{251} emphasize accurate contact modeling, joint limits, and articulated constraints. PyBullet~\cite{93, 139, 254, 263, 302, 185} remains one of the most widely used open-source engines owing to its strong community support.
MuJoCo, which is one of the most popular simulators for contact-rich tasks for its inherent contact modeling and fast simulation speed.
It powers standard benchmarks such as RoboSuite \cite{zhu2020robosuite} and Safe-Control-Gym \cite{yuan2022safe}, and is extensively used in contact-rich manipulation tasks including insertion \cite{85, 104, 279}, assembly \cite{99, 174}, and surface interaction \cite{66, 151}. 
Furthermore, it supports research in safe exploration and control \cite{29, 65, 91, 129}, as well as data collection and skill learning \cite{18, 37, 74, 77, 178}. 
It is also widely adopted in non-prehensile and articulated object manipulation \cite{7, 81, 94, 145}, dynamic and locomotion tasks \cite{100, 101}, and complex media interaction \cite{92, 106}. 
Additional applications span diverse contact-rich scenarios \cite{108, 109, 119, 120, 125, 126, 128, 133, 135, 150, 159, 170, 173, 190, 197}.

\emph{\textbf{GPU engines}} leverage massive parallelism to accelerate data collection for RL. 
The Isaac ecosystem is widely adopted, including Isaac Gym and Isaac Sim, along with their modular frameworks like Factory \cite{narang2022factory}, Orbit \cite{mittal2023orbit}, Isaac Lab \cite{mittal2025isaaclab}, and RoboVerse \cite{geng2025roboverse}. 
These platforms support large-scale training for contact-rich tasks \cite{noseworthy2025forge, 22, 207, 240, 252, 298}.
Recently, MuJoCo MJX \cite{zakka2025mujoco} has also introduced GPU-accelerated physics to the MuJoCo ecosystem.



\textbf{Domain-specific environments} support safety-critical perception and planning. Assistive Gym~\cite{93} focuses on assistive robotics scenarios, while Box2D~\cite{184} remains useful for rapid prototyping of 2D contact and control tasks.

\textbf{Interfaces and experiment factories} standardize and streamline the development of contact-rich robotic applications. OpenAI Gym-style wrappers~\cite{263, 267} provide unified observation and action APIs across simulation engines and tasks, whereas SAI~2.0~\cite{186} and related software stacks offer perception--control scaffolding for contact-rich manipulation.

\textbf{Benchmarks} are essential for quantifying progress in safety, generalization, and contact robustness. We categorize them into general manipulation suites, contact-intensive benchmarks, and safety-specific environments, as detailed in Table~\ref{tab:benchmarks}.

\textbf{General manipulation suites} provide standardized platforms for evaluating manipulation algorithms across a diverse range of tasks. RoboSuite~\cite{zhu2020robosuite} standardizes robot models and canonical tasks (e.g., insertion and door opening), serving as a primary testbed for modular algorithm design. Meta-World~\cite{yu2020meta} and RLBench~\cite{james2020rlbench} offer broad multi-task evaluation, although they often rely on simplified contact physics.
Recently, RoboVerse \cite{geng2025roboverse} and RoboCasa \cite{nasiriany2024robocasa} have scaled these to large-scale, scene-level manipulation tasks using foundation models.

\textbf{Contact-intensive and generalizable benchmarks} address the reality gap in contact physics by providing more realistic contact-rich manipulation scenarios. ManiSkill~\cite{mu2maniskill} and ManiSkill2~\cite{gumaniskill2} feature diverse articulated objects and high-fidelity physics for learning generalizable policies, including contact-rich tasks such as charger insertion and peg insertion. Factory~\cite{narang2022factory} provides physically accurate assembly tasks, such as nut-and-bolt threading, specifically designed for evaluating contact-rich policies in Isaac Gym.

\textbf{Safety-oriented benchmarks} focus explicitly on evaluating safety-aware learning and control algorithms. Safe-Control-Gym~\cite{yuan2022safe} and its successor, Safety Gymnasium~\cite{ray2019benchmarking, 152}, are the de facto standards for safe reinforcement learning, defining constrained navigation and manipulation tasks with hazard maps and cost signals. Beyond penalty-based safety, Safe-Control-Gym~\cite{yuan2022safe} integrates control-theoretic constraints, including symbolic priors and partial observability, for evaluating methods with formal safety guarantees. Robust Gymnasium~\cite{gu2025robust} further targets safety under distributional shifts and adversarial disturbances, which are critical for robust sim-to-real transfer.

\begin{tcolorbox}[
  colback=gray!3,
  colframe=gray!50,
  title={\Large\textcolor{black}{\textbf{Takeaway: Choosing Simulation and Benchmark Platforms}}},
  fonttitle=\bfseries,
  left=1mm, right=1mm, top=1mm, bottom=1mm
]
{
Engine choice is governed by a contact-fidelity-versus-parallelism trade-off: MuJoCo-family engines are favored for contact-rich fidelity, whereas GPU engines (the Isaac ecosystem) are favored for large-scale parallel data collection. More importantly for this survey, safety-specific suites (Safety Gymnasium, Safe-Control-Gym, Robust Gymnasium) remain few and largely abstract: they under-represent deformables, multi-stage and tool-supported tasks, and critically standardized contact-force evaluation, the very gap our Reporting recommendations are intended to mitigate.
}
\end{tcolorbox}

\subsection{Safety Evaluation Metrics}\label{subsec:metrics}


Safe learning works differ from traditional learning works in their primary goals. Traditionally, the learned models are evaluated for their predictive performance or their task completion success. 
Some works also focus on the efficiency of the learning system and generalization to novel situations. 
Although the safe learning works are still concerned with these metrics, they add new ones related to safety, such as a statistic of constraint violations or a cost value. 
In the case of contact-rich problems, the safety metrics are usually related to the contact forces.

\begin{figure*}[ht]
    \centering
    \includegraphics[width=0.92\linewidth]{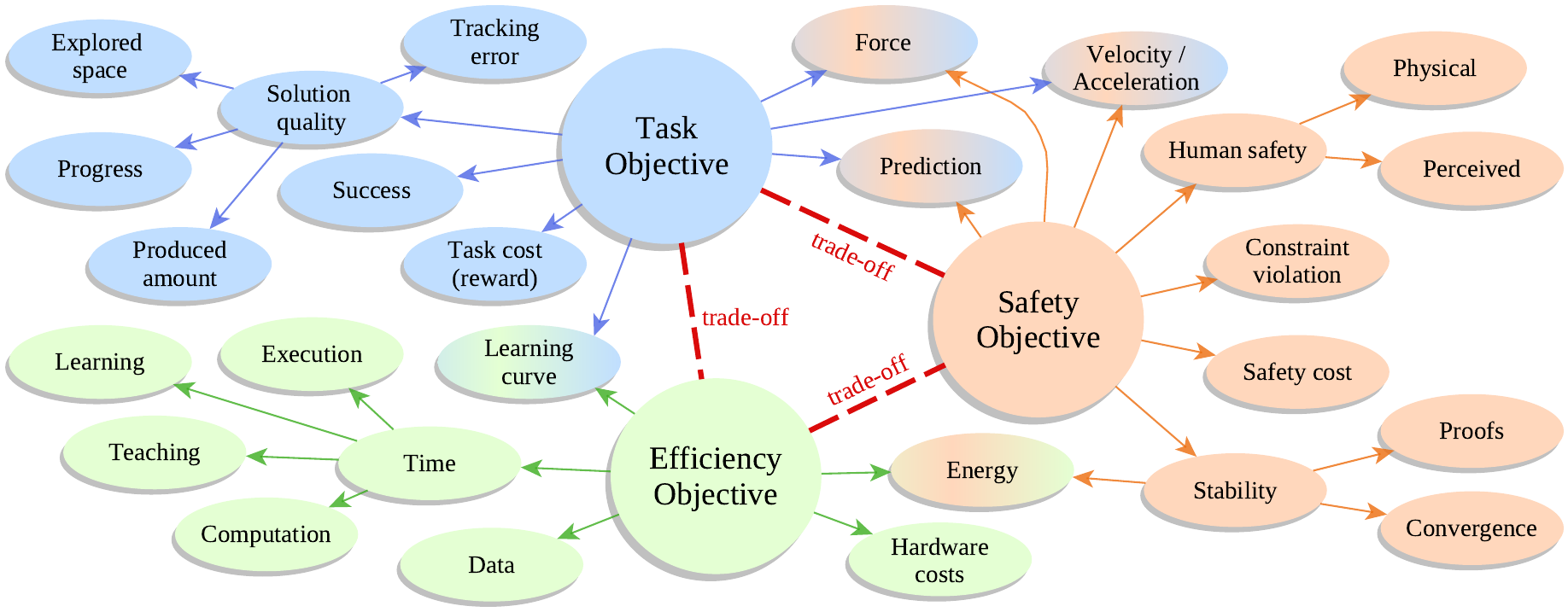}
    \caption{The metrics in the contact-rich safe learning literature can be categorized according to their objectives: Safety, task and efficiency. There are trade-offs between these objectives. 
    The researchers aim to develop safe systems with a good trade-off for task success and efficiency.
    Some metrics, such as force, velocity and energy serve to measure multiple objectives.}
    \label{fig:metrics}
\end{figure*}

In the following, we present and categorize the evaluation metrics used by the relevant works scanned in our survey. 
We present the prominent metrics as categorized according to their goals in Fig.~\ref{fig:metrics}. 
The identified metrics aim to measure the safety, task (success) and efficiency objectives. 

\textbf{Safety, efficiency, and task objectives} are the primary criteria used to evaluate safe learning methods for contact-rich manipulation.\label{sec:metrics-objectives} Safety is usually assessed through violations of predefined constraints~\citep{13, 11} or the value of a cost function~\citep{65, 331}. In physical interaction tasks, it is often quantified using physical metrics such as contact force, acceleration, and velocity~\citep{36, 44}. Control stability and human safety constitute two important subcategories of the safety objective. Stability can be evaluated through convergence rates~\citep{21, 2} or the energy flow of the system~\citep{zanella2024learning, zhang2025passivity, 375}.
Human safety includes primarily the \textit{physical safety} of the human in direct contact tasks like massaging or bathing assistance \citep{19, 5}, however, \textit{perceived safety} concepts such as trust, predictability and comfort gain importance for the acceptance of robots in social spaces \citep{52}. 
A relevant work \citep{9} evaluates the perceived safety through a 7-point Likert survey in their dressing assistance experiments.

The safety objective is the main topic of this survey, however, it cannot be considered in isolation from the other criteria. 
The task objective is quantified by the frequency of achieving success \citep{94} or the value of a task cost function (i.e., reward) \citep{95}. In some works, the quality of a solution is assessed by the progress of task completion (\cite{9}: dressing rate, \cite{21}: vegetable slicing progress), size of the explored space \citep{1}, amount of produced output (\cite{12}: ground fine powder), or the tracking accuracy \citep{5}. \cite{20} defines an instability index to quantify the episode reward deviation between runs. 
{On the other hand,} the efficiency objective is assessed by the energy consumption \citep{86}, and time requirements of the data collection (e.g., teaching) \citep{82}, learning a model \citep{99}, task execution \citep{95, 3}, or computation of an action or a plan \citep{94, 15}. 
Since robot hardware is costly, reducing real-world trials by learning in simulation is preferable \citep{4}.

\textbf{Trade-offs between safety, efficiency, and task objectives} are inherent to contact-rich robot learning because these objectives often conflict (Fig.~\ref{fig:metrics}). Safety evaluation should therefore not rely on a single metric in isolation. For example, a method that achieves a high task success rate by repeatedly exceeding contact-force limits cannot be considered truly safe, whereas a method that rarely violates safety constraints but rarely completes the task may be impractical for real-world deployment.
Having both task and safety objectives increases the optimization complexity, and consequently the amount of data to learn effective policies \citep{331}.
{Thus, in contact-rich tasks, it is expected that a safe method may achieve a lower raw success rate or longer training times than an unconstrained or unsafe baseline. What matters is whether the loss is justified by a clear reduction in unsafe contacts, excessive forces, instability, or hardware risk.}

The trade-off can be measured by a combined metric, such as the ratio of the safe and successful experiments to constraint violating ones \citep{91, zhang2024srl}.
A safety-efficiency reward function was proposed by \cite{28} to achieve a good trade-off. 
On the other hand, 
\cite{11}, \cite{7} and \cite{13} compare the numbers of successful and violating runs side-by-side instead of directly calculating a ratio between them. {Side-by-side comparison can capture more nuanced conclusions and is thus recommended when reporting the final results. However, when evaluating many runs, such as in parameter optimization and ablation studies, it is more practical to use a combined metric for direct comparison.}

{Care should be taken when reporting combined trade-off metrics. If task success, safety, and efficiency are collapsed into a single reward or cost value, it becomes difficult to determine whether a lower score is caused by poor task execution, conservative safety behavior, excessive contact forces, or inefficient motion. We therefore recommend building trade-off analyses on simple and directly interpretable quantities, such as the number of successful episodes, the number of violating episodes, and the number of safe successful episodes.}



Machine learning metrics such as prediction accuracy have been used to measure task objectives, however, they are also relevant in works that use learned models for safety-related predictions \citep{21, 4}.
Just as the learning curves provide insights on the success and efficiency trade-off, some works present the learning curves of success, violation and their ratio to portray the trade-off between the three objectives during training \citep{91, zhang2024srl}.
{We further recommend reporting variability across random seeds and training runs, especially for reinforcement learning methods, where inter-seed variation and hyperparameter sensitivity can be large. Reporting only the best run or a single seed can obscure unsafe failure modes and overstate robustness.}


Although shifting the focus to safety can decrease the task performance or learning efficiency, these objectives can be aligned in the case of physically interpretable metrics such as force and motion tracking errors, and energy consumption.
Physically meaningful results such as the force-velocity behavior can provide intuition towards the safety of the system \citep{1, 56}. 
Tracking the target force without large deviations in the Lyapunov-based boundaries directly translates to both the control stability and task achievement quality \citep{2}. 
On the other hand, learning energy-based stability from data directly translates to energy efficiency in deployment \citep{zhang2025passivity}.

\textbf{Improved safety evaluation} can be achieved by assessing system robustness under noise and disturbances.\label{sec:metrics-improved} The safety of a learning system can be further evaluated by introducing external disturbances or uncertainty during testing. For example,~\cite{5} evaluates the convergence of a trajectory-tracking controller under external disturbances. Noise can be injected into the observation space~\citep{noseworthy2025forge, 18, 94} or the action space~\citep{20}, while vision-based systems can be challenged using varying backgrounds or cluttered scenes~\citep{329}. These evaluations help reveal the limitations of learning systems and characterize their sensing and robustness requirements.
\cite{87} evaluate their system under uniformly sampled simulation parameters, in addition to localization uncertainties. \cite{337} compares the safety and success of different approaches in non-stationary environments. These and similar evaluation efforts allow demonstrating the robustness and generalization capacity of a method.

\begin{figure}[t]
    \centering
    \includegraphics[width=0.85\columnwidth]{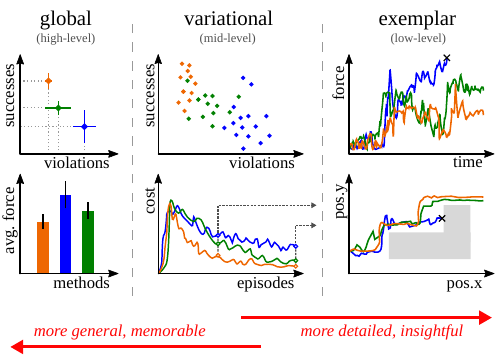}
    \caption{Three levels of result analyses: Global (high-level), variational (mid-level), and exemplar (low-level).}
    \label{fig:analysis-levels}
    \vspace{-0.3cm}
\end{figure}

Apart from the added conditions, the evaluation can be strengthened by presenting metrics at different levels of detail. 
In Fig.~\ref{fig:analysis-levels},
we define 3 levels of analyses to establish a common understanding and encourage more insightful results: \begin{enumerate*}
    \item \textit{global} (high-level) analyses present aggregated summary statistics such as the maximum force, total number of violations, or mean ($\pm$ variance) tracking error over many episodes. 
    \item \textit{variational} (mid-level) analyses show the trends of metrics among many episodes or seeds, such as a learning curve, a histogram of execution times
    , or a scatter plot of success ratio versus force violation
    .
    \item \textit{exemplar} (low-level) analyses present descriptive case studies, often real-time measurements from a single episode, such as motion trajectories, force measurements, or even multimedia material.
\end{enumerate*}
Going downwards, the amount of detail and depth of analysis increase; going upwards, the results get more general and memorable. 
\textit{Variational} comparisons allow a more in-depth analysis of the system behavior from temporal, spatial or multi-variate perspectives. 
\textit{Exemplar} results can give valuable insights on why and how the method works, or when it fails, without deriving general conclusions. 
We recommend providing analyses at all levels for a more comprehensive communication of the results. 

\textbf{Reporting recommendations} begin with presenting a compact yet interpretable set of evaluation metrics rather than relying on a single reward or cost.\label{sec:reporting-recommendations} At a minimum, evaluations of safe learning for contact-rich tasks should report task success, constraint violations, violation severity, and efficiency, together with variability across random seeds or repeated runs. Violation severity should further be complemented by physically meaningful contact metrics, such as maximum or mean contact force, force-tracking error, or injected energy when applicable.
For learning-based methods, the evaluation should also report the parameter-tuning protocol, including the number of seeds and the search range of important hyperparameters. 
Since hyperparameter choices and tuning seeds can substantially affect final performance and sample efficiency, separating tuning and testing conditions improves the fairness and reproducibility of comparisons~\citep{eimer2023hyperparameters}. 
When possible, the same metrics should also be reported under controlled perturbations or distribution shifts. This reflects the practical reality of contact-rich deployment, where pose, sensing, friction, stiffness, geometry, and human motion can vary across trials. In this way, safety evaluation can communicate not only whether a method works, but also how it trades off success, safety, efficiency, robustness, and tuning effort.

\subsection{Safety Abstraction Level}\label{sec:abstraction-level}
In contact-rich robotic tasks, safety can be embedded at various abstraction levels in the system hierarchy, ranging from high-level task planning to low-level control, and even spanning across end-to-end learning architectures. Each level presents different trade-offs between interpretability, reactivity, and generalizability. This section categorizes learning-based safety approaches by the level of abstraction at which they operate and discusses their respective advantages, challenges, and applications.
{This abstraction-level view is orthogonal to the \emph{enforcement-space} view 
: the present section asks \emph{at which layer} of the hierarchy a safety mechanism is introduced (planning, control, or end-to-end), whereas 
\emph{enforcement-space}
asks \emph{in which coordinate space} the resulting constraints are expressed and enforced (task, joint, or policy space). A given method occupies one position on each axis.}

\textbf{High-level planning} typically encodes safety through symbolic reasoning, task graphs, or decision-making policies that govern long-horizon robot behavior.
Learning-based approaches at this level often involve hierarchical reinforcement learning (HRL) \cite{239, 365}, safe policy search with task constraints \cite{251, 47}, or logic-based planning augmented by learned models \cite{118, 311}.
In contact-rich settings, these methods aim to avoid unsafe task sequences or infeasible contact configurations by reasoning over constraints such as force thresholds, joint limits, or task success probabilities \cite{204, 208}.

For example, shielded planning mechanisms and arbitration graphs can learn to prune unsafe high-level actions using safety classifiers or predictive models \cite{banerjee2024dynamic, 227, 292}. 
While such methods provide good interpretability and modularity, they depend heavily on accurate abstractions and may suffer from limited reactivity to unforeseen contact dynamics during execution.
Temporal logic specifications have also been integrated into diffusion-based planning to satisfy temporally-extended symbolic constraints \cite{218}.

Recently, the emergence of LLMs and VLMs has enabled advanced high-level planning capabilities for robotic systems, including in contact-rich tasks. 
These models can interpret complex task instructions, reason about safety constraints from natural language or visual cues, and generate high-level action sequences or code that proactively avoid unsafe situations \cite{8, 307, 330, 115}. 
For example, VLM-enhanced planners can leverage multimodal perception to identify hazardous regions or delicate objects in the environment, integrating this information into symbolic task planning or policy generation \cite{33, 51, 309, 360}. 
This allows robots to adapt their strategies dynamically, ensuring that safety considerations are embedded from the outset of task execution.

\textbf{Low-level control} ensures safe interactions at the actuation and feedback levels by enforcing physical constraints in real time. These constraints include limits on force, torque, impedance, and stability, which are particularly critical in contact-rich scenarios.
Learning-based methods here typically involve adaptive impedance control, learning CBFs, model-predictive safety filters, or robust policy adaptation techniques
\citep{30}.

{Compliant control and impedance learning} constitute one of the most widely adopted strategies for achieving safe low-level control in contact-rich manipulation. Instead of learning direct torque commands, these methods learn variable impedance or admittance parameters~\citep{23, 25, 28, 317}. By modulating stiffness and damping online according to contact feedback, they ensure compliant behavior that naturally limits interaction forces and absorbs impacts~\citep{1, 40, 74}. Recent approaches further integrate passivity constraints or energy tanks into the learning process to guarantee coupled stability during dynamic contact~\citep{zhang2025passivity, 35}.

{Safety filters and barrier-based methods} provide hard safety guarantees by wrapping a learned policy with a protective control layer.\label{sec:lowlevel-filter} To enforce constraints such as joint limits, friction cones, and maximum contact forces, control barrier functions (CBFs) and related safety filters monitor the control input and minimally modify it to keep the system within a forward-invariant safe set~\citep{29, 235}. Model Predictive Safety Filters (MPSFs) extend this idea by reasoning over a finite prediction horizon to prevent inevitable future violations, which is particularly important for systems with inertia or actuation limits~\citep{21, banerjee2024dynamic}.

{Robust and adaptive safety} addresses uncertainties arising from imperfect models and changing task conditions. For scenarios involving unknown friction, payload variations, or other modeling uncertainties, robust control techniques such as sliding mode control (SMC) and adaptive control are combined with learning to preserve safety margins~\citep{5, 60, 94}. These low-level controllers compensate for disturbances and unmodeled dynamics that might otherwise destabilize a high-level learned policy.

\textbf{End-to-end safety enhancement} seeks to integrate perception, decision-making, and actuation within a unified model while embedding safety either implicitly (e.g., through regularization) or explicitly (e.g., via constrained optimization or safety critics).\label{sec:e2e_vla}
This includes safe reinforcement learning frameworks that learn both task policies and safety constraints simultaneously from raw sensory data \cite{287, 99, 101}.
For instance, \cite{100} proposed a neural network-based approach that learns to predict safe actions directly from raw sensory inputs, such as images and proprioceptive data, allowing the robot to adapt its behavior in real-time to avoid unsafe actions during contact-rich tasks.

Recently, VLA models have emerged as a powerful paradigm for end-to-end safety enhancement. 
These models bridge perception, reasoning, and control by learning from large-scale multimodal datasets that include language instructions, visual observations, and robot actions \cite{330, 33, 10705419}.
Safety is enhanced in VLA systems through several mechanisms:
(i) Safety-aware prompting and grounding: VLA models can be conditioned on safety-specific prompts or rules (e.g., ``move slowly near the glass") to generate compliant behaviors \cite{8, 51}.
(ii) Safety critics and filtering: Auxiliary safety modules or critics can be trained to evaluate the risk of generated actions before execution, filtering out potentially hazardous commands \cite{376, 360, 115}.
(iii) Multimodal alignment: By aligning visual and linguistic representations with physical safety constraints (e.g., force limits), VLA models can better generalize safety concepts to novel objects and environments \cite{309, 307, 218}.
While these methods promise scalability and generalization, especially in contact-uncertain environments, they often suffer from limited interpretability and pose challenges in formal safety verification compared to modular architectures.

{To make this concrete, Fig.~\ref{fig:ppe_stack} instantiates the prevailing {plan--parameterize--enforce} (PPE) architecture of recent safe VLA systems with named modules. In the \emph{plan} stage, a VLM/VLA parses the instruction and scene into a task sketch and a set of explicit, machine-checkable safety rules (e.g., a force cap on a cable, a no-touch label on a painted face, a speed cap near a human), optionally grounding physical reachability as in PhysVLM~\citep{zhou_physvlm_2025}. 
In the parameterize stage, mid-level skills expose safety--relevant references, including target pose, force, and impedance or stiffness gains, so the semantic plan is compiled into bounded, compliant skill parameters rather than raw torques~\citep{376, lu_vla-rl_2025}. In the \emph{enforce} stage, a low-level shield (a CBF-QP, a reachability filter, or an MPC safety filter) projects each command onto the admissible set at run time, bounding force, region, and velocity regardless of upstream errors~\citep{330}. Two complementary design choices recur: SafeVLA casts safety as a \emph{constrained-learning} objective, so the policy is trained to respect explicit cost limits rather than relying on prompt compliance alone~\citep{33}, whereas tactile-augmented models (Tactile-VLA~\citep{huang2025tactile}, VTLA~\citep{zhang2025vtla}, BiTLA~\citep{yang2025bitla}, OmniVTLA~\citep{cheng2025omnivtla}) add contact/force tokens to the observation so the policy can condition on, and react to, contact onset and incipient slip rather than vision alone.}

{This layering matters because semantic and physical safety can diverge. Consider a ``wipe the table but do not press the cup'' instruction in which the VLM mis-grounds the fragile-object mask and treats the cup as part of the table surface. At the plan level the action remains semantically valid, wiping is permitted, so prompt- and alignment-based checks pass; the error surfaces only physically, as the end-effector drives its tracking setpoint into the cup and a contact-force spike develops. In a PPE stack the enforce layer intercepts this: a force/region control barrier function detects the barrier value $h(\boldsymbol{x})\!\rightarrow\!0$ at the no-touch boundary and the force limit, minimally modifies the command to arrest penetration before an overload occurs, and logs the event for recalibration. The same failure would go undetected by a language-only safeguard. This illustrates why language-level constraint specification alone is insufficient for physical contact safety, a point we revisit among the VLM/VLA-specific challenges and why a certified low-level shield remains the component that actually bounds interaction forces.}

\begin{figure}[t]
\centering
\begin{tikzpicture}[
  node distance=5.5mm,
  box/.style={draw=black, line width=0.6pt, rounded corners, align=center, inner sep=4pt, text width=0.84\columnwidth, font=\footnotesize},
  ar/.style={-{Latex}, draw=black, line width=0.7pt}
]
\node[box] (plan) {\textbf{Plan} (VLM/VLA): parse instruction $+$ scene $\rightarrow$ task sketch and explicit, machine-checkable safety rules (force caps, no-touch regions, speed limits)};
\node[box, below=of plan] (param) {\textbf{Parameterize} (mid-level skills): compile the plan into pose / force / impedance references rather than raw torques};
\node[box, below=of param] (enforce) {\textbf{Enforce} (low-level shield): CBF-QP / reachability / MPC filter projects each command onto the admissible set and logs violations};
\node[box, below=of enforce] (robot) {\textbf{Robot} $+$ environment contact (force / tactile / pose feedback to the shield)};
\draw[ar] (plan) -- (param);
\draw[ar] (param) -- (enforce);
\draw[ar] (enforce) -- (robot);
\end{tikzpicture}
\caption{{The plan--parameterize--enforce (PPE) architecture used by recent safe VLA systems, instantiated with named modules. The plan stage (e.g., SafeVLA~\citep{33}; safe robot foundation models~\citep{330}) emits a task sketch and explicit safety rules; the parameterize stage exposes pose/force/impedance references; the enforce stage applies a certified CBF/reachability/MPC shield. Tactile-augmented variants (e.g., Tactile-VLA~\citep{huang2025tactile}) feed contact/force signals back to the shield, which is the component that actually bounds interaction forces when upstream semantic grounding is wrong.}}
\label{fig:ppe_stack}
\end{figure}

{Crucially, these foundation-model mechanisms are not new safety \emph{principles} but re-instantiations, at the semantic level, of the classical notions.
Language- and rule-to-constraint grounding is the foundation-model counterpart of constraint satisfaction and reachability;
safety critics, refusal, and risk-aware filtering of generated actions instantiate risk- and data-driven safety;
the low-level shields beneath a VLA planner are exactly the invariance/CBF certificates;
conservative default force/velocity caps and compliant references reduce to the force/energy/passivity view;
no-touch and human-proximity rules map to human-centered and standard-driven safety;
and the grounding of intent itself is the semantic-safety axis.
Table~\ref{tab:fm_principles} summarizes this correspondence. Read this way, the contribution of robotic foundation models to contact-rich safety is to \emph{specify and ground} constraints from open-ended language and perception, whereas the guarantee that those constraints actually hold during contact still rests on the classical low-level mechanisms surveyed above. The open problem is therefore not to replace those mechanisms but to ensure that the semantic layer feeds them correct, physically meaningful constraints, a requirement that current systems meet only partially.}

\begin{table}[t]
    \centering
    {
    \footnotesize
    \renewcommand{\arraystretch}{1.2}
    \caption{{Mapping from foundation-model (VLM/VLA) safety mechanisms to the classical safety principles they instantiate 
    }}
    \label{tab:fm_principles}
    \begin{tabularx}{\columnwidth}{@{}>{\raggedright\arraybackslash}p{0.40\columnwidth} >{\raggedright\arraybackslash}X@{}}
    \toprule
    \textbf{Foundation-model mechanism} & \textbf{Classical safety principle it instantiates}\\
    \midrule
    Grounding language and rules into formal constraints & Constraint satisfaction \& reachability \\
    Semantic scene grounding (affordances, no-touch, fragile regions) & Semantic safety \\
    Safety critics, refusal, risk-aware action filtering & Risk- \& data-driven safety \\
    Low-level shields under a VLA planner (CBF / reachability / MPC) & Stability \& set invariance \\
    Conservative default caps and compliant references & Physical-interaction safety: force, energy, passivity \\ 
    Human-proximity / speed–separation / PPE rules & Human-centered \& standard-driven safety \\
    \bottomrule
    \end{tabularx}
    }
\end{table}

\textbf{Hybrid methods} combine safety mechanisms across multiple abstraction levels to leverage the strengths of each.\label{sec:hybrid-methods} For instance, a high-level planner may filter task options using a safety-aware graph while a low-level controller enforces contact-force constraints through learned CBFs~\citep{9531394}. Alternatively, end-to-end models may be guided by high-level logic rules or shielded by low-level safety filters during deployment.

{Plan--learn--filter pipelines} represent one of the most common hybrid architectures for safe contact-rich manipulation. A high-level planner (graph, sampling, optimization, or semantic decomposition) proposes nominal contact strategies or intermediate sub-goals; a learned policy (often variable impedance/admittance or residual) generates parameterized actions (pose, force, or stiffness); and a runtime safety layer (CBF, reachability, MPC shield, or action projection) minimally modifies unsafe commands while preserving task progress~\citep{25, 28, 29, 26, martinez2015safe, yuan2022safe}. This layered design reduces unsafe transients during approach, impact, and sustained contact by deferring exact constraint satisfaction to an online safety filter.

{Compliant action abstraction and certified projection} provide another effective hybrid strategy by learning safe action representations instead of directly predicting joint torques. Rather than outputting raw torques, these methods learn action abstractions such as desired Cartesian displacements, force targets, or diagonal/structured stiffness gains that preserve passivity margins while exposing explicit interfaces for safety modification~\citep{zhang2025passivity, zhang2024srl, 25, 28, 317, roveda2020model}. Certified projection, implemented through quadratic programs with CBF constraints, shield vetoes, or reachability pre-checks, enforces forward invariance or constraint satisfaction without retraining the underlying policy~\citep{29, 26, yuan2022safe, 152}.

{Generative or residual policies under safety wrappers} combine expressive policy representations with certified runtime safety mechanisms. Diffusion or residual reinforcement learning policies generate candidate motions or compliance updates, while a safety critic or barrier/MPC layer evaluates predicted force envelopes, energy flow, or constraint margins before execution ("generate--score--project")~\citep{100, zhang2024srl, 29, zhang2025passivity, 362, 117, 12}. Residual admittance and stiffness adaptation compensate for unmodeled contact geometry, whereas the surrounding safety wrapper prevents excessive contact forces and other unsafe behaviors~\citep{25, 28, 317, 178}.

{Semantic or VLM/VLA hybrid safety} extends hybrid architectures by integrating vision-language(-action) models into the safety pipeline. Vision-language(-action) and other foundation-model components parse unstructured instructions and scene context into structured constraint representations~\citep{33, 330, 10705419}. These constraints are subsequently passed to the plan--learn--filter stack, where a semantic planner produces task goals or action sequences, a learned variable-impedance or residual policy generates compliant control parameters~\citep{376, 360}, runtime CBF, reachability, or MPC safety layers enforce physical limits, and a logging module records violations for calibration and continual improvement~\citep{25, 28, 29, yuan2022safe}. Multimodal fusion, particularly of vision, tactile sensing, and force sensing, further disambiguates occlusions and anticipates risky contact transitions, while structured prompting and action schemas reduce hallucination-induced unsafe commands~\citep{100, 33, 330}. This hybridization aligns high-level semantic intent with certifiable low-level execution, enabling scalable yet physically grounded safe manipulation in contact-rich settings.

Such multi-layered architectures can offer robustness and flexibility, especially in unstructured contact-rich tasks. However, hybrid methods also introduce complexity in integration and may require careful coordination to avoid conflicts between safety modules operating at different levels.

\subsection{Safety Enforcement Spaces}\label{sec:enforcement-space}

In the context of learning for contact-rich manipulation, safety is not only a monolithic concept but also is enforced hierarchically across the robot's control architecture. Constraints can be imposed at the level of physical interaction (task space), internal actuation (joint space), or within the decision-making abstraction (policy space).
Each of these domains provides distinct opportunities and challenges for incorporating safety. Task-space enforcement directly regulates end-effector motion and contact forces with the environment, joint-space methods shape internal robot behavior to avoid unsafe configurations or torques, while policy-level mechanisms ensure that the agent’s decision-making respects safety constraints from the outset. This section categorizes existing literature based on these architectural levels, synthesizing the dominant control strategies and emerging hybrid paradigms.
{This enforcement-space view is complementary to, not a repetition of, the abstraction-level view:
there we asked at which layer safety is added, here we ask in which coordinate space the constraints live. To avoid duplication, the cross-layer \emph{plan--learn--filter} hybrids are treated once, under \emph{Hybrid methods};
the present section instead focuses on how constraints expressed in task, joint, and policy space differ and combine.}

\textbf{Task-space safety mechanisms} operate directly at the interface of physical interaction by regulating the kinematics and dynamics of the end-effector. By defining safety in Cartesian coordinates, these methods directly constrain the contact forces and motions exerted on the environment.
A predominant strategy for task-space safety is the use of compliance and stiffness modulation. Rather than tracking rigid trajectories, impedance and admittance control schemes allow the robot to deviate from reference paths in response to external forces, thereby absorbing impact energy and maintaining stable contact \citep{4, 6, 7, 12, 16}. The works in \cite{1, 3} explicitly bound the forces and torques at the end-effector to prevent excessive contact pressures, integrating such constraints into the controller and learning process. Recent learning-based frameworks have extended this architecture by parameterizing variable impedance policies, where the agent learns to output context-aware stiffness or virtual force targets to ensure gentle interaction \citep{23, 25, 107}. To guarantee stability in these interactions, energy-based approaches often enforce passivity constraints, ensuring that the robot behaves as a passive system and does not inject unbounded energy into the environment \citep{35, 75}.

Beyond inherent compliance, formal constraints are frequently applied to bound the workspace. CBFs and safety filters are employed to strictly confine end-effector states within a safe set, overriding policy actions that threaten to violate positional or force limits \citep{14, 57, 80}. These safe manifolds can be learned from demonstrations or modeled via Gaussian processes, restricting the planner to regions with high probability of safety \citep{21, 101}.
In reinforcement learning settings, task-space safety is often formulated via reward shaping to penalize excessive contact forces \citep{noseworthy2025forge, 20, 62, 84}, or by restricting the action space to a safe subset of the workspace to prevent collisions during exploration\citep{11, 22, 96, 98}. The works in \citep{8, 34} also use planning-based approaches to build safety into motion generation. Before execution or learning, they compute contact-rich trajectories that satisfy contact stability, task unrelated collision avoidance, and joint limits, so that any subsequent policy learning begins with inherently safe trajectories.

Notably, in human–robot interaction scenarios, the constraints are adapted to dynamic environments, utilizing safety fields to enforce separation distances, and controllers are designed to comply with industry safety standards, e.g. ISO force limits, during physical contact \citep{15, 51, 93}.

\textbf{Joint-space safety mechanisms} prioritize the robot's intrinsic integrity, whereas task-space methods focus on extrinsic interaction. These methods operate in the configuration space, constraining joint angles, velocities, and torques to prevent self-collisions, singularity proximity, and actuator saturation.

Safety in this domain is often realized by augmenting low-level controllers with limiting functions. Joint-level CBFs and saturation filters ensure that commanded torques do not exceed hardware capabilities or result in unsafe kinetic energy, thereby indirectly limiting the force transfer to the environment \citep{44}. Stability is further ensured through Lyapunov-based formulations, such as enforcing positive-definiteness in learned joint stiffness matrices to guarantee system passivity \citep{53, 59}.

To mitigate the risks of dynamic impacts, adaptive joint-torque controllers utilize proprioceptive feedback to dissipate shock through joint compliance \citep{60, 94}. In the context of RL, policies operating in joint space are frequently constrained to prevent erratic high-frequency actuation. Strategies include penalizing proximity to joint limits \citep{97} or employing hierarchical schemes where the RL agent dictates high-level goals while a lower-level motion planner generates smooth, collision-free configuration trajectories \citep{99}. Specific applications, such as robotic massage, have demonstrated that limiting joint-space velocities and applying compliance controls effectively bounds the output torque, ensuring safe physical interaction through intrinsic actuation limits \citep{46}.

\textbf{Dual-space safety enforcement} combines task-space and joint-space constraints to eliminate the safety blind spots associated with relying on either representation alone. While task-space control may inadvertently command infeasible joint configurations, joint-space safety mechanisms may overlook excessive contact forces or pressures at the end-effector. Consequently, hybrid architectures have emerged to enforce safety concurrently across both spaces, achieving a balance between safe physical interaction and protection of the robot itself.
{This dual-space sense of \emph{hybrid} is distinct from the cross-\emph{level} hybrids:
here it refers to enforcing constraints simultaneously in two \emph{coordinate spaces} (task and joint), not to combining abstraction levels (planning, control, end-to-end). We therefore keep the plan--learn--filter discussion in \emph{Hybrid methods}
and restrict this subsection to task/joint coordination.}

Hierarchical control frameworks typically employ a task-space loop to manage environmental interaction forces, coupled with an inner joint-space loop to handle redundancy resolution and torque limits. A Cartesian impedance controller regulates contact stiffness, while a secondary joint-space reflex module monitors for collisions or internal limits, overriding commands to dissipate energy if discrepancies arise \citep{66, wang2022high}. Theoretical guarantees for such dual-loop architectures have been established via Lyapunov analysis, proving that errors in both force tracking and joint deviation remain bounded \citep{56, michel2021bilateral}. Additionally, in \cite{liu2025safe}, the constrained manifold is imposed in both joint and Cartesian space to ensure safety.

Ensuring stability in the joint space can indirectly enforce task-space safety \citep{long2025neural}. In \cite{88}, a model-based actor–critic method confines the joint controller’s behavior within stability limits using an analytical model of the robot and environment stiffness. As a result, the end-effector also never pushes beyond safe force limits. Similarly, in dynamic locomotion tasks like hopping, strictly respecting joint torque and angle limits is shown to inherently stabilize the task-space center-of-mass trajectory, preventing falls \citep{101}.

\textbf{Policy-space safety enforcement} incorporates safety directly into high-level decision making by embedding safety considerations into the learning objective or planning horizon. Rather than relying solely on real-time control interventions, these methods proactively avoid states or action sequences that would otherwise require low-level emergency corrections.

A standard formalism for this is the CMDP, where the agent optimizes a reward signal subject to a cost threshold associated with unsafe states \citep{agnihotri2024cop, ding2023provably, wang2023enforcing, yu2022towards}. Lagrangian relaxation is commonly used to dynamically weight this cost, training the policy to satisfy safety constraints asymptotically \cite{33}. Alternatively, safety can be encoded into the value function itself\citep{165, tan2024safe}. Safety critics or feasibility value functions estimate the long-term probability of constraint violation, guiding the policy to select actions that maximize the margin of safety \citep{45}. To handle potential failures, \citep{91} introduces a separate safety policy that intervenes when the agent approaches a high-risk region, steering the system back to a safe state before returning control to the task policy. 

Implementing such boundaries becomes non-trivial when the state space is high-dimensional or unstructured. In high-dimensional visual manipulation tasks, safety barriers are often learned in latent spaces, allowing the agent to distinguish between safe and unsafe visual manifolds without explicit state estimation \citep{65, tan2024safe, wilcox2022ls3}. Furthermore, when safety constraints are difficult to formalize mathematically, Human-in-the-Loop paradigms incorporate human oversight to shape the policy's risk sensitivity, utilizing human feedback and intervention to define safety boundaries during the exploration phase \citep{54}. Beyond defining constraints and boundaries, generative models such as diffusion policies have been employed to synthesize action sequences that prioritize smoothness and low uncertainty, thereby reducing the stochasticity that often leads to unsafe behaviors \citep{42, guodon, zhang2025constrained}.

{We summarize how the task-, joint-, dual-, and policy-space enforcement strategies differ along unified criteria in Table~\ref{tab:space_comparison}.}

\begin{sidewaystable}[t]
      \centering
      {
      \footnotesize
      \renewcommand{\arraystretch}{1.25}
      \caption{{Structured comparison of safety \emph{enforcement spaces} for contact-rich learning, organized by coordinate space. For each space we summarize what is constrained and where safety lives, the
  resulting guarantee type, the modeling and training assumptions it relies on, its compute placement and cost, and its characteristic blind spots. The supporting references for each space are given in the text
  of this section.}}
      \label{tab:space_comparison}
      \newcolumntype{G}[1]{>{\hsize=#1\hsize\RaggedRight\arraybackslash}X}%
      \begin{tabularx}{\textwidth}{@{}l G{0.90} G{1.0} G{1.0} G{1.0} G{1.0}@{}}
      \toprule
      \textbf{Enforcement space} & \textbf{Where safety lives} & \textbf{Guarantee type} & \textbf{Assumptions} & \textbf{Compute placement \& cost} & \textbf{Limitations}\\
      \midrule
      Task space & Cartesian end-effector motion and contact force/wrench at the physical interface, workspace region & Direct contact-force/region bounding via compliance/impedance \& passivity, or
  forward-invariant safe set via CBF/safety filter, can be hard on the EE state when the model holds & Accurate task-space kinematics and an impedance/force controller, known force/region limits, reliable wrench
  sensing & Online QP/safety filter or compliant controller at the control rate, modest per-step cost & May command kinematically infeasible or limit-violating joint configurations, ignores internal actuator
  and self-collision integrity\\
      Joint space & Configuration-space quantities, joint angles, velocities, torques, actuator saturation, singularity proximity, self-collision & Intrinsic integrity via joint limits and
  saturation filters plus Lyapunov/passivity & Accurate robot dynamics/actuator model, proprioceptive feedback, access
  to low-level torque control & Low-level torque/saturation filter or joint-level CBF at the servo rate, lightweight & May overlook excessive end-effector contact pressure at the environment interface, safe
  joints do not imply safe interaction\\
      Dual-space & Both coordinate frames concurrently, EE force/motion {and} joint limits/torques & Combined: task-space invariance/compliance with
  joint-limit and Lyapunov guarantees, bounded-error proofs for both force tracking and joint deviation & Both a task-space interaction model and a joint-space dynamics model, consistent redundancy resolution,
  dual-loop coordination & Coupled dual loop controllers, higher integration and coordination overhead & Integration complexity and possible inter-loop conflicts, larger modeling
  burden than either single space\\ 
      Policy space & Decision-level quantities, CMDP cost/value, expected constraint cost, feasibility/safety value, shapes exploration before low-level intervention & Typically soft/asymptotic, constraint
  satisfaction in expectation via CMDP/Lagrangian relaxation or feasibility value functions / safety critics, preemptive rather than instantaneous & Adequate training coverage, a specified cost threshold/labels, state representation, frequently a safe backup policy & Training-time constrained optimization with lightweight runtime policy/critic evaluation & Only preemptive/asymptotic, no instantaneous
  hard bound, needs a low-level backup or shield, degrades under distribution shift and high-dimensional state\\
      \bottomrule
      \end{tabularx}
      }
  \end{sidewaystable}

 \subsection{Current Limitations and Research Gaps} \label{sec:challenges}
    Despite rapid progress in learning-based methods for contact-rich robotic manipulation, several \textbf{key challenges} remain before these approaches can be deployed safely and reliably at scale. This section outlines major open problems, with particular attention to how emerging VLM/VLA-based methods both amplify and help address these issues.

    \textbf{Fragile safety and stability guarantees} remain a fundamental challenge for learning-based controllers, including those using deep networks and VLAs, which often lack formal guarantees of passivity, stability, and constraint satisfaction under hard contacts~\citep{zhang2025passivity, spyrakos2020passivity, 75}. This limitation can lead to unsafe behaviors and hardware damage. While passivity-centric and energy-constrained formulations offer promising directions, integrating them seamlessly with large, expressive policies remains technically challenging, especially when contact conditions change rapidly or unpredictably.

\textbf{Data scarcity and biased supervision} continue to hinder the development of robust contact-rich manipulation systems. Contact-rich skills require precise force control and tight tolerances, but high-quality demonstrations and interaction data are expensive, brittle, and difficult to scale compared to standard vision--language datasets. VLMs and VLAs promise better generalization through pretraining, yet they are typically supervised by weak or indirect signals that do not capture fine-grained contact events, leading to policies that may look competent visually but fail under subtle force or geometry variations.

\textbf{Limited robustness, generalization, and uncertainty awareness} remain major obstacles to reliable deployment. Learning-based policies for contact-rich manipulation, especially those trained on narrow task distributions, often overfit to specific layouts, materials, or sensing conditions and degrade sharply under distribution shifts. Although VLAs offer broader prior knowledge, current models typically have weak uncertainty estimation, making it difficult to detect when a contact-rich interaction~\cite{33, 369} is entering an unsafe regime and requiring fallback to safer behaviors or human intervention.

    \textbf{Safe exploration under physical risk} remains the most safety-critical phase, as naive probing can cause jamming, surface damage, or tool wear in insertion, polishing, and assembly contexts~\citep{yuan2022safe, zhang2024srl}. Existing constrained and risk-sensitive reinforcement learning frameworks provide expectation- or probabilistic-based safety guarantees, yet remain vulnerable under distribution shifts. Shielding and action-projection mechanisms, including control barrier functions, reachability analysis, and model predictive safety filters, provide hard online guarantees but typically rely on conservative uncertainty sets, thereby limiting exploration efficiency and data utilization.

\textbf{Safe execution and performance trade-offs} remain an equally important challenge after policy learning. Ensuring sustained constraint satisfaction during deployment requires balancing responsiveness, task efficiency, and physical safety margins. Variable impedance or admittance policies learned from demonstrations or reinforcement signals can adapt compliance online~\citep{28, 25, zhang2024srl}; however, their stability margins under rapid parameter adaptation are rarely quantified. Moreover, runtime safety filters may degrade task performance through overly conservative interventions, whereas softer filtering strategies risk cumulative force or constraint violations. Existing methods therefore lack a principled framework for jointly optimizing (i) safety violation probability, (ii) force and energy envelopes, and (iii) task efficiency.
    
     
    \textbf{Benchmark and dataset limitations} continue to hinder the systematic evaluation of safe learning methods for contact-rich manipulation. Existing simulation suites and benchmarks provide limited coverage of deformable objects, multi-stage processes, tool-supported tasks, and tasks performed in close collaboration with humans~\citep{towers2024gymnasium, geng2025roboverse, gu2025robust}. Public datasets rarely include (i) high-frequency wrench traces aligned with failure annotations; (ii) rich near-miss episodes essential for learning recovery policies; or (iii) semantic safety labels, such as forbidden regions and fragile surfaces. This scarcity impedes reproducible comparisons of safety violations, recovery latency, and generalization. Although emerging safety-oriented RL frameworks and robust benchmarks partially address constrained exploration, they still lack standardized protocols for evaluating contact-force safety~\citep{152, raffin2021stable}. Until standardized safe-contact benchmarks mature, future evaluations can follow the reporting recommendations, which emphasize interpretable task, safety, efficiency, variability, and robustness metrics.

\textbf{Safety risks specific to VLMs and VLAs} arise from the introduction of generalist, language-conditioned policies into contact-rich manipulation. Such models may misinterpret language instructions, hallucinate feasible contact strategies, or overconfidently execute plans that violate implicit safety constraints. Recent work on safety-aligned VLAs and failure detection for generalist policies highlights the difficulty of systematically eliciting unsafe behaviors, constraining them during learning, and verifying that deployment remains within acceptable risk bounds. A specific consequence of conflating semantic and physical safety
concerns the interpretation of \emph{evidence}. Language-level constraint specification, such as prompting ``do not exceed 5 N'' or ``do not touch the painted face,'' demonstrably improves semantic or plan-level behavior, but to our knowledge no published study has isolated language-level specification and shown that it alone reduces \emph{measured} contact-force or impulse violations on real hardware. Reported safety gains are typically evaluated at the planning or refusal level, or they rely on downstream low-level controllers that enforce physical constraints regardless of the language input. We therefore caution against interpreting semantic-level safety as evidence of physical contact safety, and identify the controlled evaluation of this relationship, namely, whether language-level constraints reduce real contact-force violations when the low-level safety layer is held fixed or ablated, as an important open research question and a concrete benchmark target.

\textbf{Human--robot collaboration safety} introduces additional challenges beyond physical constraint satisfaction by requiring robots to account for both hard safety limits (e.g., force and speed) and softer human-centered factors such as comfort, predictability, and trust. During contact-rich interaction, operators must be able to anticipate robot behavior and intervene effectively, yet current learning-based controllers often lack transparency and interpretable failure modes. Existing safe learning pipelines predominantly encode physical thresholds while overlooking temporal predictability and trust metrics that are central to user acceptance~\citep{346, 253, hanna2020towards}. Although adaptive safety frameworks have begun integrating ergonomic considerations and standard-driven constraints, such as the force envelopes defined in ISO/TS 15066~\citep{ISO15066}, their combination with learning-enabled adaptation remains limited. Advanced models, including VLAs, can in principle explain their intentions or request human assistance; however, \revthree{designing effective interaction protocols, shared-autonomy mechanisms, and explanations that are both faithful and useful in high-contact scenarios remains an important open challenge.}

    \subsection{Perspectives}\label{sec:perspectives}
    Safe robot learning for contact-rich manipulation is entering a phase where systematic advances in data, evaluation, and architectural composability must complement incremental improvements in algorithms. We outline key forward-looking directions.
    
    \textbf{Standardized safe-contact benchmarks} are essential for advancing reproducible evaluation of safe learning methods. Current benchmark suites under-represent multi-stage forceful interactions, deformable objects, and human-adjacent tasks~\citep{towers2024gymnasium, geng2025roboverse, gu2025robust, yuan2022safe}. Future benchmarks should move beyond isolated manipulation tasks to standardized suites that stress high-force interactions, tight clearances, and diverse materials and fixtures. A key opportunity is to design benchmarks in which VLM/VLA models must interpret natural-language task descriptions and safety rules before executing contact-rich skills under explicit constraints on contact forces, slippage, and failure recovery. Future benchmark suites should also adopt consistent reporting practices, such as those summarized in Safety Evaluation Metrics section, to facilitate fair comparison of safety, task performance, efficiency, and robustness across methods. In particular, benchmarks should make it possible to attribute safety gains to the appropriate system layer, for example, by fixing or ablating the low-level safety filter to evaluate whether language-level constraint specification alone reduces measured contact-force violations. 

\textbf{Safety-aware and data-efficient learning frameworks} will be central to the next generation of contact-rich robotic systems. Future methods should tightly integrate formal safety mechanisms with advanced function approximators, including VLAs and other large policies pretrained on internet-scale multimodal data and robot demonstrations. Promising directions include risk-sensitive and constraint-aware VLA controllers, hybrid model-based/model-free frameworks that combine high-level reasoning from large models with certified low-level safety filters, and approaches that leverage language for specifying, verifying, and adapting safety constraints during contact-rich interaction.

\textbf{Bridging the sim-to-real safety gap} remains a prerequisite for deploying safe learning algorithms in real-world environments. Closing this gap will require simulators that faithfully reproduce contact dynamics while also capturing the perceptual and linguistic variability encountered by VLM/VLA policies. This opens the possibility of using generative models and large-scale VLAs to create diverse synthetic curricula of tasks, objects, and safety-critical edge cases, followed by safety-aware adaptation and online refinement on physical robots with continuous monitoring of contact forces, constraint violations, and failure modes.

\textbf{Multimodal semantic grounding} is another critical direction for improving the reliability of VLA-based manipulation. Although VLA models have demonstrated impressive capabilities in scene understanding and semantic reasoning~\citep{33, 330, 10705419}, robust grounding remains fragile under occlusion, clutter, and ambiguous physical interactions. A promising direction is layered grounding, progressing from scene parsing to affordance and constraint extraction, followed by formal constraint compilation (e.g., CBF/MPC formulations) and ultimately runtime conformance validation. In addition, safety-aware prompting and structured action representations can help reduce the risk of hallucinated unsafe behaviors~\citep{geminiroboticsteam2025evaluatinggeminiroboticspolicies}.

    
    \textbf{Human-centric and lifecycle-aware safety} extends beyond enforcing physical safety envelopes, such as limits on force and speed, which are necessary but not sufficient for the widespread adoption of collaborative robots~\citep{ajoudani2018progress}. Future safe learning frameworks can benefit from language-conditioned policies that understand instructions, preferences, and corrections expressed in natural language, enabling non-expert users to shape safe behavior during contact-rich collaboration. Over the system lifecycle, advanced models such as VLAs may also support continual learning and self-diagnostics, allowing robots to explain their contact strategies, identify emerging risks, and adapt their policies while preserving guarantees of human safety and equipment integrity.

\textbf{Scalable verification for learned controllers} remains a major challenge as modern policy representations become increasingly expressive. Diffusion policies, latent-variable policies, and residual impedance controllers are difficult to analyze using classical verification techniques. A promising direction is to abstract policy outputs, such as stiffness and damping parameters, into reachable or invariant sets that admit formal safety analysis. Likewise, data-driven certificates, including learned control barrier functions and neural safety critics, require systematic mechanisms for failure detection, online adaptation, and certificate repair to prevent the gradual degradation of safety guarantees during long-term deployment.

\textbf{Physically-aware foundation models} offer another promising direction for improving safe contact-rich manipulation. Current generalist manipulation models often fail to capture the underlying mechanics of physical interaction, resulting in unsafe extrapolation beyond the training distribution. Integrating structured physical priors, such as friction cones, passivity constraints, energy tanks, and contact mechanics, into large-scale action models could substantially improve physical reliability. A hybrid ``plan--parameterize--enforce'' paradigm, in which a foundation model generates high-level plans that are subsequently refined under impedance and barrier-based safety constraints, provides a principled framework for combining semantic reasoning with certified low-level execution~\citep{100, 330}. Furthermore, building large-scale datasets that are rich in contact interactions and explicitly annotated with safety-relevant events, failure modes, and physical constraints will be essential for training reliable next-generation robotic foundation models.
    
   \textbf{Open problems} remain at the intersection of learning, control, and physical interaction. Key unresolved issues include certified generalization under simultaneous geometric and material shifts~\citep{das2025towards}, the synthesis of minimally conservative safe sets for multi-contact interactions, robust grounding of semantic constraints into physically enforceable safety guarantees, and lifelong adaptation with formal transient safety bounds. Addressing these challenges will require tighter integration of analytical safety frameworks with data-driven representations and robotic foundation models, enabling systems that are both adaptive and verifiably safe under complex contact dynamics.

\textbf{The outlook for safe learning in contact-rich robotics} is increasingly shaped by the convergence of physically grounded control, formal safety theory, and foundation-model-based intelligence. Future progress will depend on combining physically grounded control methods, such as impedance control and passivity, with formal safety tools including control barrier functions and reachability analysis, scalable data resources such as standardized benchmarks, and high-level semantic reasoning enabled by VLMs and VLAs. A growing consensus supports layered ``plan--learn--filter--execute'' architectures that integrate semantic planning, adaptive learning, and certified low-level safety enforcement into a unified framework. Ultimately, establishing standardized evaluation protocols, interpretable safety metrics, and reproducible benchmarks will be essential for translating recent advances in safe learning into reliable deployment of autonomous robotic systems operating in complex, contact-rich environments.

\section{Discussion}
In this survey, a comprehensive overview of safe learning methods for contact-rich robotic tasks has been presented, with a particular focus on how emerging foundation models reshape the landscape. Existing approaches have been organized along multiple dimensions, including sensing modalities, data acquisition strategies, simulation environments, safety evaluation metrics, abstraction levels, and task mobility contexts, to provide a structured view of the design space for safe contact-rich manipulation.

The survey highlights the central role of multimodal sensing, such as force/torque, tactile, and vision, in enabling safe and robust behavior under complex contact conditions. Key challenges persist in data acquisition, safety metrics and certification, modeling of contact-rich interactions, and consistently enforcing safety constraints from high-level reasoning down to low-level control. Although we systematically searched multiple major academic databases and applied predefined inclusion criteria, the literature search was restricted to studies published between 2018 and 2025 and may not capture all relevant work, particularly rapidly emerging research on robotic foundation models and safe contact-rich learning.

Looking forward, hybrid architectures and advanced learning paradigms offer promising avenues. Opportunities include combining high-level planning with low-level compliant control, integrating classical safety mechanisms with learning-based policies, and leveraging VLM and VLA models to interpret safety requirements, ground them in multimodal observations, and support transparent, human-centered interaction in contact-rich scenarios.


\section{Methods}
To ensure a comprehensive and systematic review of safe learning methods for robotic contact-rich tasks, we employed a multi-step methodology: 
     \subsection{\revthree{Literature Investigation}} We conducted an extensive investigation across major academic databases, including IEEE Xplore, ACM Digital Library, SpringerLink, and Google Scholar. Keywords used in the literature search included combinations of "safe", "contact-rich tasks," and "learning". To capture the most recent advancements, specifically, we use the following compliant search string: \textit{("robot" OR "robotic") AND ("safe" OR "safety" OR "stable") AND ("contact-rich" OR "multi-contact" OR "contact-intensive" OR "physical interaction" OR "environment interaction" OR "contact interaction" OR "contact manipulation" OR "contact force") AND (intitle:"learning" OR intitle:"learn")}. {The database search was last executed in August 2025; in addition, we screened the reference lists of the related surveys discussed above for further relevant works (backward snowballing).
     \subsection{\revthree{Inclusion Criteria}} We focused on both peer-reviewed and preprint articles published from 2018 to 2025 that specifically addressed and/or discussed safety in learning-based approaches for contact-rich robotic tasks. Both theoretical contributions and empirical studies with experimental validation were considered.
     \subsection{Analysis} Each paper was analyzed for its methodological contributions, safety guarantees and discussion among multiple different perspectives, experimental setups including simulation and real-world experiments, and limitations. 
     \subsection{\revthree{Exclusion Criteria}}\label{sec:scope}
     We excluded works that (i) do not involve contact-rich interaction as defined in the section on Robotic Contact-Rich Tasks (e.g., free-space motion or pure pick-and-place), (ii) do not incorporate any learning component, (iii) do not address safety, stability, or constraint satisfaction in any form, or (iv) fall outside the 2018--2025 window. Duplicate records and non-archival materials were also removed.
     \subsection{\revthree{Selection and Screening}} Records were processed in the standard identification--screening--eligibility--inclusion stages summarized in the PRISMA-style diagram of Fig.~\ref{fig:prisma}. Title/abstract screening and full-text eligibility assessment were each performed independently by two of the authors, and disagreements were resolved by discussion. The stage-wise counts, including records identified, records after de-duplication, records screened, full texts assessed, records excluded with reasons, and studies finally included, are reported in Fig.~\ref{fig:prisma}.
\begin{figure}[t]
\centering
\begin{tikzpicture}[
  font=\scriptsize,
  main/.style={draw=black, line width=0.6pt, rounded corners, align=left, inner sep=4pt, text width=4.8cm},
  excl/.style={draw=black, line width=0.5pt, rounded corners, align=left, inner sep=3pt, text width=3.0cm},
  ar/.style={-{Latex}, draw=black, line width=0.6pt},
  node distance=5mm
]
\node[main] (id) {\textbf{Identification.} Records identified via database search (IEEE Xplore, ACM DL, SpringerLink, Google Scholar), 2018--2025: $n=6,472$};
\node[main, below=of id] (dedup) {\textbf{De-duplication.} Records after duplicates removed: $n=4,850$};
\node[main, below=of dedup] (screen) {\textbf{Screening.} Title/abstract screened: $n=4,850$};

\node[excl, right=3mm of screen, yshift=12mm] (screxc) {Excluded at screening: $n=3,920$};

\node[main, below=of screen] (elig) {\textbf{Eligibility.} Full texts assessed: $n=930$};
\node[excl, right=3mm of elig] (eligexc) {Excluded with reasons (not contact-rich; no learning component; no safety aspect; outside 2018--2025): $n=538$};
\node[main, below=of elig] (inc) {\textbf{Included.} Studies included in the review: $n=392$};

\draw[ar] (screen.east) to[out=0, in=270] (screxc.south);

\draw[ar] (id) -- (dedup);
\draw[ar] (dedup) -- (screen);
\draw[ar] (screen) -- (elig);
\draw[ar] (elig) -- (inc);
\draw[ar] (elig.east) -- (eligexc.west);

\end{tikzpicture}
\caption{{PRISMA-style flow of the systematic review, with the identification, de-duplication, screening, eligibility, and inclusion stages; the databases and the exact Boolean search string are given in the text.}}
\label{fig:prisma}
\end{figure}

\section{Declarations}

\subsection{\revthree{Data Availability}}
The curated literature dataset used to organize and analyse the studies included in this review is publicly available in the project repository at:
\url{https://github.com/jack-sherman01/Awesome-Learning4Safe-Contact-rich-tasks/tree/main/docs}
\subsection{Code Availability}
The scripts used to curate and analyse the literature included in this review are available in the project repository at \url{https://github.com/jack-sherman01/Awesome-Learning4Safe-Contact-rich-tasks/tree/main/scripts}. We also share the paper collection list with links for readers, which is available explicitly at:
\url{https://github.com/jack-sherman01/Awesome-Learning4Safe-Contact-rich-tasks}

\subsection{Acknowledgements}
This work was supported by the European Union Horizon Project TORNADO (Grant GA 101189557).
\subsection{\revthree{Author Contributions}}
HZ conceived the study and led the manuscript development. HZ, RD, GS, and PZ conducted the literature investigation and analysis and prepared the initial draft of the manuscript. NT, YS, and AA contributed to the interpretation of the literature and provided critical revisions to the manuscript. AA supervised the study. All authors revised the manuscript, approved the final version, and agree to be accountable for the work.

\subsection{Competing Interests}
The authors declare no competing interests.
\bibliography{reference_V3}

\end{document}